\documentclass[10pt,twocolumn,letterpaper]{article}

\usepackage{cvpr}              % To produce the CAMERA-READY version

\usepackage[accsupp]{axessibility}  % Improves PDF readability for those with disabilities.

\usepackage[ruled, linesnumbered]{algorithm2e}
\usepackage{algpseudocode}
\usepackage{caption}
\usepackage{subcaption}
\usepackage{graphicx}
\usepackage{amsmath}
\usepackage{amssymb}
\usepackage{booktabs}
\usepackage{bm}
\usepackage{enumitem}
\usepackage{multirow}
\usepackage{makecell,rotating}
\usepackage[normalem]{ulem}
\newcolumntype{L}[1]{>{\raggedright\arraybackslash}p{#1}}
\newcolumntype{C}[1]{>{\centering\arraybackslash}p{#1}}
\newcolumntype{R}[1]{>{\raggedleft\arraybackslash}p{#1}}

\usepackage[pagebackref,breaklinks,colorlinks]{hyperref}

\usepackage[capitalize]{cleveref}
\crefname{section}{Sec.}{Secs.}
\Crefname{section}{Section}{Sections}
\Crefname{table}{Table}{Tables}
\crefname{table}{Tab.}{Tabs.}

\def\cvprPaperID{4118} % *** Enter the CVPR Paper ID here
\def\confName{CVPR}
\def\confYear{2023}

\begin{document}

% \title{\LaTeX\ Author Guidelines for \confName~Proceedings}
% \title{Multi-Constrained Layout Generation with Layout Constrained Decoding}
\title{LayoutFormer++: Conditional Graphic Layout Generation via Constraint Serialization and Decoding Space Restriction}
% \author{
% Zhaoyun Jiang\\
% Xi'an Jiaotong University\\
% {\tt\small jzy124@stu.xjtu.edu.cn}
% \and
% Jiaqi Guo\\
% Microsoft Research Asia\\
% % First line of institution2 address\\
% {\tt\small jiaqiguo@microsoft.com}
% \and
% Shizhao Sun\\
% Microsoft Research Asia\\
% % First line of institution2 address\\
% {\tt\small shizhao.sun@microsoft.com}
% \and
% Huayu Deng\\
% Shanghai Jiaotong University\\
% % First line of institution2 address\\
% {\tt\small deng\_hy99@sjtu.edu.cn}
% \and
% Zhongkai Wu\\
% Beihang University\\
% % First line of institution2 address\\
% {\tt\small 17376487@buaa.edu.cn}
% \and
% Vuksan Mijovic\\
% Microsoft\\
% % First line of institution2 address\\
% {\tt\small vmijovic@microsoft.com}
% \and
% Zijiang James Yang\\
% Xi'an Jiaotong University\\
% % First line of institution2 address\\
% {\tt\small zijiang@xjtu.edu.cn}
% \and
% Jian-Guang Lou\\
% Microsoft Research Asia\\
% % First line of institution2 address\\
% {\tt\small jlou@microsoft.com}
% \and
% Dongmei Zhang\\
% Microsoft Research Asia\\
% % First line of institution2 address\\
% {\tt\small dongmeiz@microsoft.com}
% }
\author{
  Zhaoyun Jiang\textsuperscript{\rm 1}\thanks{Work done during an internship at Microsoft Research Asia.\protect \label{fn}} , 
  Jiaqi Guo\textsuperscript{\rm 2}, 
  Shizhao Sun\textsuperscript{\rm 2}, 
  Huayu Deng\textsuperscript{\rm 3}\textsuperscript{\ref*{fn}},
  Zhongkai Wu\textsuperscript{\rm 4}\textsuperscript{\ref*{fn}}, \\
  Vuksan Mijovic\textsuperscript{\rm 5},
  Zijiang James Yang\textsuperscript{\rm 1},
  Jian-Guang Lou\textsuperscript{\rm 2},
  Dongmei Zhang\textsuperscript{\rm 2} \\
  \textsuperscript{\rm 1}Xi'an Jiaotong University,
  \textsuperscript{\rm 2}Microsoft Research Asia, \\
  \textsuperscript{\rm 3}Shanghai Jiaotong University,
  \textsuperscript{\rm 4}Beihang University,
  \textsuperscript{\rm 5}Microsoft,  \\
  {\tt\small jzy124@stu.xjtu.edu.cn, deng\_hy99@sjtu.edu.cn, 17376487@buaa.edu.cn,} \\
  {\tt\small zijiang@xjtu.edu.cn, \{jiaqiguo, shizsu, vmijovic, jlou, dongmeiz\}@microsoft.com}
}
\maketitle

%%%%%%%%% ABSTRACT
\begin{abstract}
    \label{sec:abstract}
% The automatically generated layouts should meet both quality and control requirements. Although existing works have made meaningful attempts in improving the generation quality, they failed in achieving the control requirements that model and exert multiple user constraints. In this work, we propose [Approach Name] for constrained layout generation, which can flexibly generate layouts from diverse user constraints and effectively avoid the constraint violations. To achieve these, we first propose an identical sequence format for various input constraints. We leverage an encoder-decoder architecture with Transformers for [Approach Name] to process the constraints modeling and the layout prediction respectively. To impose the constraints on the generated layout, we introduce [Constrained decoding strategy], forcing the constraints by pruning the infeasible values according to the given constraints during the decoding process. Experiments on two public datasets and six layout generation tasks with different user constraints demonstrate that [Approach Name] significantly outperforms the previous approaches, achieves both quality and control requirements for the constrained layout generation.

Conditional graphic layout generation, which generates realistic layouts according to user constraints, is a challenging task that has not been well-studied yet.
First, there is limited discussion about how to handle diverse user constraints flexibly and uniformly.
Second, to make the layouts conform to user constraints, existing work often sacrifices generation quality significantly.
In this work, we propose LayoutFormer++ to tackle the above problems.
First, to flexibly handle diverse constraints, we propose a constraint serialization scheme, which represents different user constraints as sequences of tokens with a predefined format.
Then, we formulate conditional layout generation as a sequence-to-sequence transformation, and leverage encoder-decoder framework with Transformer as the basic architecture.
Furthermore, to make the layout better meet user requirements without harming quality, we propose a decoding space restriction strategy.
Specifically, we prune the predicted distribution by ignoring the options that definitely violate user constraints and likely result in low-quality layouts, and make the model samples from the restricted distribution.
Experiments demonstrate that LayoutFormer++ outperforms existing approaches on all the tasks in terms of both better generation quality and less constraint violation.

\end{abstract}

%%%%%%%%% BODY TEXT
\section{Introduction}
\label{sec:introduction}

\begin{figure}[t]
    \begin{center}
    \includegraphics[width=\linewidth]{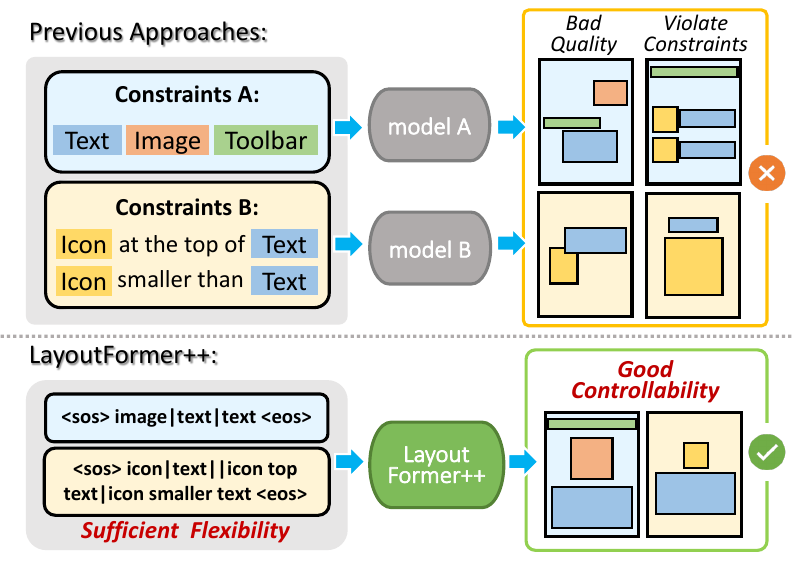}
    \end{center}
    \setlength{\abovecaptionskip}{-0.1cm}   
    \setlength{\belowcaptionskip}{-0.45cm}
    \caption{Comparing with previous conditional layout generation approaches, LayoutFormer++ performs better on sufficient flexibility and good controllability.}
    \label{fig:conditional_layout_generation}
\end{figure}

Graphic designs greatly facilitate information communication in our daily life. 
During its creation, the \emph{layout}, i.e., positions and sizes of elements, plays a critical role. 
To assist layout design, \emph{conditional layout generation}, which takes user constraints as input and generates layouts as output, attracts great attention (see Figure~\ref{fig:conditional_layout_generation}).
It is different from unconditional layout generation, which generates layouts freely without constraints, from at least two aspects.
First, the model should be able to handle diverse user constraints, called \emph{sufficient flexibility}. 
Figure~\ref{fig:layout_generation_tasks} shows 6 typical tasks of layout generation in real-world applications including layout completion, layout refinement, layout generation conditioned on element types, element types with sizes, element relationships, or any of their combinations.
Second, the model should generate layouts conforming to user requirements (i.e., constraints) as many as possible without harming quality, called \emph{good controllability} (see Figure~\ref{fig:conditional_layout_generation}).

% \vspace{-0.4cm}  

%While existing work makes meaningful attempts, conditional layout generation is still far from being well-studied.
However, existing work cannot meet the above two requirements. 
%First, there is limited discussion about how to achieve sufficient flexibility.
First, there is no existing work can support all the layout generation tasks with different user constraints.
Most existing approaches simply focus on tackling a single conditional layout generation task without considering whether they can be applied to other tasks.
For example, LayoutTransformer~\cite{LayoutTransformer} can only perform completion task and BLT~\cite{BLT} cannot handle element relationships. Such task-specific approaches hinder the development of solutions for the new task.
Second, there are no satisfactory methods to ensure good controllability.
Some approaches directly replace the values in the predicted layout with the ones specified in the user constraints~\cite{LayoutTransformer,BLT}.
Another work defines a set of heuristic rules and leverages latent optimization~\cite{LayoutGAN++}.
To make the generated layouts meet user requirements, they often degrade the generation quality significantly.

In this work, we propose a unified model called \emph{LayoutFormer++} to support the different scenarios of conditional layout generation.
In \emph{LayoutFormer++}, a set of constraints is represented as a sequence.
Specifically, we use a \emph{constraint serialization} scheme to serialize different user constraints into a sequence of tokens with a predefined format (see Figure~\ref{fig:conditional_layout_generation}).
The intuition behind this design choice is as follows.
%On the one hand, the sequence format is expressive and inclusive, which has been successfully used in typical CV~\cite{} and NLP tasks~\cite{}.
First, the sequence format is widely used and effective.
Its effectiveness in layout generation has also been demonstrated in recent works~\cite{LayoutTransformer,BLT}.
%On the other hand, the constraints for layout generation are suitable to be represented as sequences by nature.
Second, a sequence is very flexible to accommodate different constraints for layout generation. 
We can serialize any structured information in the user constraints as a sequence.
We found although user needs are diverse, they are all about element types and five attributes including type, top coordinate, left coordinate, width and height.
Thus, we can simply define a set of vocabularies to describe the attributes respectively and concatenate descriptions of different attributes and elements to construct a sequence.

Therefore, the conditional layout generation problem can be formulated as a sequence-to-sequence transformation problem.
This enables us to leverage a simple yet effective encoder-decoder framework with Transformer~\cite{Transformer} as a basic model architecture.
The encoder processes the user constraints in a bidirectional way.
The decoder predicts the layout sequence autoregressively, where there are multiple decoding steps and the model samples one token from the predicted distribution at each decoding step.

Furthermore, to achieve good controllability, we introduce a \emph{decoding space restriction} strategy in the inference stage.
Our key idea is to prune the infeasible options in the predicted distribution and make the model sample from the restricted distribution at each decoding step.
Specifically, we leverage two kinds of information to prune the options.
First, the options that definitely violate the user constraints are pruned.
For example, if a user wants one image and two buttons, the option for putting one text box will not be acceptable. 
Second, the options with low probabilities in the predicted distribution, which will very likely result in low-quality layouts, are also pruned.
As the feasible option set may be empty after pruning, we further introduce a backtracking mechanism, in which the model goes back to a certain decoding step and find a better solution.
Note that the whole generation process of the proposed strategy still relies on the distribution learned from the training data.
Thus, it is less likely to disturb a layout when making it better conform to user constraints.

% To evaluate \emph{LayoutFormer++} and compare it with state-of-the-art approaches, we developed a common platform for fair comparison. Based on the platform, 
We conduct extensive experiments on two public datasets~\cite{rico,publaynet} and six layout generation tasks with different user constraints, to evaluate \emph{LayoutFormer++} and compare it with state-of-the-art approaches.
% which are separately explored by previous work. 
Experimental results show that LayoutFormer++ can successfully tackle all six layout generation tasks that are handled separately by previous work, demonstrating that it is able to provide sufficient flexibility.
Furthermore, LayoutFormer++ significantly outperforms previous approaches in terms of both better generation quality and less constraint violation, indicating that it achieves good controllability.

\section{Related Work}
\label{sec:related_work}

\begin{figure}[t]
    \begin{center}
    \includegraphics[width=0.95\linewidth]{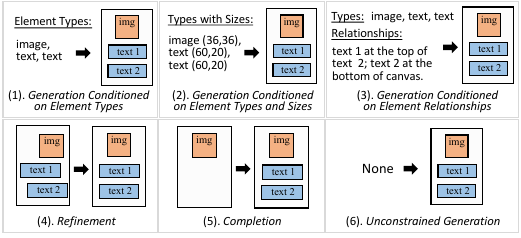}
    \end{center}
    \setlength{\abovecaptionskip}{-0.1cm}   
    \setlength{\belowcaptionskip}{-0.3cm}
    \caption{Typical tasks for conditional layout generation.}
    \label{fig:layout_generation_tasks}
\end{figure}

\textbf{Graphic Layout Generation.}
Graphic layout generation aims at generating aesthetic layouts to tackle users' needs.
As users' needs are diverse in real-world applications, various tasks of layout generation are explored (see Figure~\ref{fig:layout_generation_tasks}), such as generation condition on element types~\cite{NDN,LayoutGAN++,BLT}, generation condition on types and sizes~\cite{BLT}, generation condition on element relationships\cite{NDN,LayoutGAN++}, refinement~\cite{RUITE} and completion~\cite{LayoutTransformer}. Some other approaches focus on unconstrained generation~\cite{LayoutGAN,VTN,coarse2fine,CanvasVAE}.

However, none of existing approaches can achieve sufficient flexibility to handle diverse layout generation tasks, due to the limitations in constraint modeling. 
Some works represent the constraints in the same format of the target layout, thus can not handle other complex constraints~\cite{LayoutTransformer,BLT,RUITE}. 
For example, LayoutTransformer~\cite{LayoutTransformer} can only handle completion task but cannot deal with relationships between elements.
Other works design the modeling formats specific to the constraint, which are hard to adapt to other tasks. For example, \cite{NDN} leverages graph format to represent the relationships among the elements, which is inconvenient for the size constraints. 
In this work, we propose LayoutFormer++ with constraint serialization scheme, modeling different constraints into same format of the sequence, eliminating the isolation between layout generation tasks.

Existing approaches also consider different methods for constraint satisfaction. 
Some approaches directly reset the predicted attributes by the values specified in the constraints~\cite{LayoutTransformer,BLT}.
Other approaches define the heuristic rules and cost functions to control the generation~\cite{LayoutGAN++,NDN}. 
For example, \cite{LayoutGAN++} designs cost functions for each constraint and takes the constrained generation as an optimization problem. 
However, these methods often decrease the generation quality significantly.
For \cite{LayoutTransformer,BLT}, directly resetting the sequence will make the generated layout deviate from the distribution learned by model.
For \cite{LayoutGAN++, NDN}, it is hard to define the rules and cost functions to handle the trade-off between constraint satisfaction and generation quality.
Different from existing approaches, we introduce decoding space restriction strategy to prune the infeasible values during decoding process, which does not rely on sophisticated rules or cost functions, and also fits the learned distribution.

\textbf{Decoding Methods.}
Decoding method plays an important role in autoregressive generation. The most prominent decoding methods include Greedy Search, Beam Search, Top-K sampling and Top-P sampling~\cite{Holtzman2020The}, etc. To enable the control over the output sequence, constrained decoding is first proposed in controllable text generation. 
\cite{anderson-etal-2017-guided,hokamp-liu-2017-lexically,post-vilar-2018-fast,hu-etal-2019-improved,lu-etal-2022-neurologic,lu-etal-2021-neurologic} propose constrained decoding methods for hard lexical constraints where a set of tokens are required to occur in target output.
\cite{zhang-etal-2020-language-generation,qin2022cold} present decoding methods for soft constraints which assign a value between 0 and 1 to indicate how constraints are satisfied.
\cite{scholak-etal-2021-picard,shin-van-durme-2022-shot,krishnamurthy-etal-2017-neural} introduce decoding methods for structural constraints which impose a rigorous structures on the output sequence.
% Previous works~\cite{hokamp-liu-2017-lexically,anderson-etal-2017-guided} explore constrained beam search to exert lexical constraints, i.e. impose explicit phrase-based constraints to be placed on target output. \cite{post-vilar-2018-fast,hu-etal-2019-improved} further improve the efficiency of the constrained decoding method. 
Inspired by these methods, we propose a new constrained decoding method tailored for conditional layout generation.
%, which is inspired by the constrained decoding that restrict the prediction during the decoding process. 
We carefully design two pruning modules and a backtracking mechanism to improve the satisfaction of constraints while maintaining high layout quality.

\section{Approach}
\label{sec:approach}

\subsection{Overview}
\label{sec:approach_overview}

\begin{figure*}[htb]
    \centering
    \begin{minipage}{0.48\linewidth}
    \begin{subfigure}[Model Architecture]{\linewidth}

        \includegraphics[width=\linewidth]{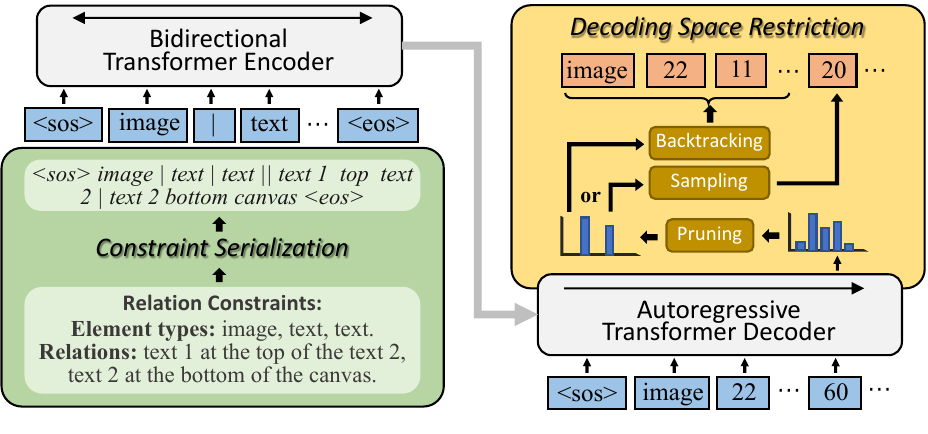}
        \caption{Model Architecture}
        \label{fig:architecture}
    \end{subfigure} %
    \end{minipage}
    \medskip
    \quad
    \quad
    \quad
    \begin{minipage}{0.36\linewidth}
    \begin{subfigure}[Input and Output Sequences]{\linewidth}   

        \includegraphics[width=\linewidth]{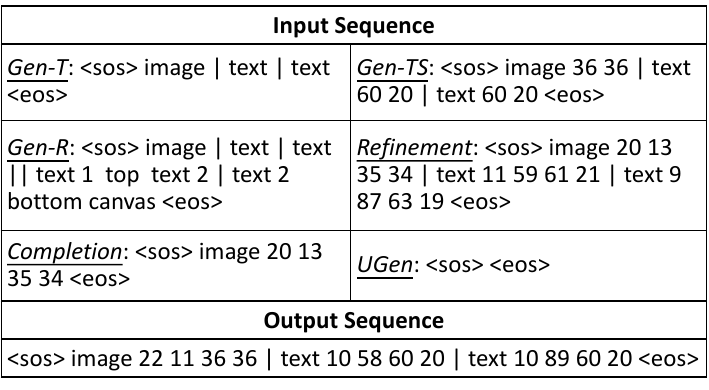}
        \caption{Input and Output Sequences}
        \label{fig:IO}    
    \end{subfigure} 
    \end{minipage}
    \setlength{\abovecaptionskip}{-0.1cm}
    \setlength{\belowcaptionskip}{-0.25cm}
    \caption{\ref{sub@fig:architecture} The overall model architecture of LayoutFormer++, illustrated by taking the task Gen-R as an example. \ref{sub@fig:IO} The examples of input and output sequences for different conditional layout generation tasks through constraint serialization.}
    \label{fig:model}
\end{figure*}

A layout is made up of a set of elements, where each element can be described by its type $c$, left and top coordinate $x$ and $y$, as well as width $w$ and height $h$. 
The continuous attributes $x$, $y$, $w$ and $h$ are quantized, which is proven to be helpful for graphic layouts~\cite{LayoutTransformer,VTN,RUITE}.
Thus, an element can be represented by five tokens.
Following the state-of-the-art approaches, we represent a layout by concatenating all the elements' tokens, i.e., $L = \{\langle\text{sos}\rangle  c_1 x_1 y_1 w_1 h_1 \dots c_N x_N y_N w_N h_N  \langle\text{eos}\rangle\}$.
Here, $N$ denotes the total number of elements, $\langle\text{sos}\rangle$ and $\langle\text{eos}\rangle$ are special tokens indicating the start and end of a sequence.

% In this work, we focus on the conditional layout generation, which takes the various user given constraints as input, and output the layout that satisfy the user constraints while achieve good quality.

% To achieve the sufficient flexibility and good controllability of conditional layout generation, we propose a new approach named LayoutFormer++, a Transformer encoder-decoder framework equipped with the constraint serialization and decoding space restriction. The overall model architecture is illustrated in Figure~\ref{fig:architecture}. 

In this work, we propose a new approach named LayoutFormer++ for conditional layout generation. The overall model architecture is illustrated in Figure~\ref{fig:architecture}.

To achieve sufficient flexibility, different constraints are formatted uniformly into sequence through constraint serialization. The details of the constraint serialization scheme will be introduced in Section~\ref{sec:constraint_serialization}.

We leverage a Transformer encoder-decoder architecture for LayoutFormer++ based on the experience of previous works.
The bidirectional encoder takes the constraint sequence as input and outputs their contextualized representations. The autoregressive decoder iteratively predicts the probabilities of future tokens. 
The model is trained to minimize the negative log-likelihood of layout tokens, i.e.,
\begin{small}
    \begin{align}\label{eqn:loss}
        \text{minimize} \sum_{t=1}^{N} -\log{P_{\bm{\theta}}(L_t | L_{<t}, S)},
    \end{align}
\end{small}where $S$ is the input constraint sequence, $L$ is the output layout sequence, $\bm{\theta}$ refers to the model parameters and $N$ denotes the length of the layout sequence.

Finally, we introduce the decoding space restriction strategy with a backtracking mechanism into inference process to achieve good controllability. 
The details of the decoding space restriction strategy are given in Section~\ref{sec:decoding_space_restriction}.

\subsection{Serializing Constraints}
\label{sec:constraint_serialization}

\subsubsection{Principles}
\label{sec:formulation_principle}
% Serializing the user constraints basically faces two problems. The first problem is how to represent each constraint in sequence format, and the second is how to combine the sequences of each constraint into a complete sequence. To solve these two problems, we introduce the constraint serialization scheme, by giving the principles in representing and combining the constraints into sequence. 
Instead of directly presenting concrete sequences for the constraints considered in previous work, we first introduce the principles for serializing constraints.
These principles not only summarize how the existing constraints are serialized but also provide a general guidance for formulating other constraints. 
There are two critical questions in serializing constraints.
First, how to represent each constraint in a sequence format.
Second, how to combine different constraints into a complete sequence.

\noindent\textbf{Constraint Representation.}
To present a constraint as a sequence, the basic idea is to first define a set of vocabularies and then concatenate the tokens from these vocabularies.
Specifically, the constraints often fall into two categories.

First, some constraints are only related to a single element. 
For example, put an image on the page or the element should be $10$ pixel wide.
The serialization for such constraint is quite straightforward: we just define a vocabulary and then represent the constraint by a single token from the vocabulary.
For example, for the constraints about element type, the vocabulary could be $\{\texttt{image}, \texttt{toolbar}, \dots, \texttt{button}\}$, and the constraint `an image' can be represented as \texttt{image}.

Second, other constraints are related to multiple elements.
For example, put an image on top of a button.
To distinguish the elements referred in the constraint, we first build a vocabulary representing unique IDs for the elements.
Then, we define a vocabulary for the relationships between elements.
At last, we concatenate the tokens of element IDs and relationships.
For example, the vocabulary for element IDs could be $\{\texttt{image}_1,\dots, \texttt{image}_{K_1},\dots,\texttt{button}_{1},\dots, \texttt{button}_{K_t}\}$, and the vocabulary for relationships could be $\{\texttt{top},\dots, \texttt{small}\}$.
Then, the constraint `an image on top of a button' 
is represented as $\texttt{image}_1 \;\; \texttt{top} \;\; \texttt{button}_1$.

% Among all the single constraints, some constraints are related to only one element, while others involve multiple elements. Based on this, we divide the single constraints into \emph{single-element constraint} and \emph{multi-element constraint}, and introduce the representation principles for them respectively.

% \begin{itemize}[itemsep=3pt,topsep=2pt,parsep=0pt,leftmargin=10pt]

%     \item \emph{Representing single-element constraint}. The single-element constraint usually specifies an attribute of one elements, e.g., assigning the type for an element. We just need to define the vocabulary to describe the corresponding attribute, and represent it by the single token.

%     \item \emph{Representing multi-element constraint}. For the multi-element constraint, we first specify an unique ID for each element by concatenating its type with an index, to distinguish the elements referred in the constraint. Then we define the vocabulary for the correlation among the elements described by the constraint. Finally, we concatenate the tokens of element IDs and their correlations, to formulate the sequence for the multi-element constraint.
    
% \end{itemize}

\noindent\textbf{Constraint Combination.}
To combine different constraints, we concatenate their sequences.
To ease the learning of neural network, we use a fixed order instead of a random order when concatenating different constraints.
First, for the constraints only related to a single element, we first concatenate those for the same element and then concatenate those from different elements.
% For example, for a constraint about types and sizes of different elements, the sequence after concatenating all the constraints is \texttt{image 20 10  | button 2 4}.
Second, for the constraints related to multiple elements, we put them after all the constraints for the single elements.

% After representing the constraints by sequences, we need to combine these sub-sequences into a complete sequence. In the following, we introduce how we connect the sequences of the single-element constraints and the multi-element constraints.

% \begin{itemize}[itemsep=3pt,topsep=2pt,parsep=0pt,leftmargin=10pt]

%     \item \emph{Combining single-element constraints}. 
%     First, we group the tokens of single-element constraints by their related elements, and concatenate tokens in the order of $c,x,y,w,h$ within the group. Then, the sequences of each group are concatenated by the alphabetic order of the related element types to formulate the complete sequence.
    
%     \item \emph{Combining multi-element constraints}. We concatenate the sequences of the multi-element constraints in the alphabetic order of the types of elements that the constraints involved.
    
% \end{itemize}
% We choose connecting order depends on the different tasks and constraints.
% According to the practical experience, we add the special tokens $\langle\text{sos}\rangle$ and $\langle\text{eos}\rangle$ at the start and the end of the sequence to indicate the start and end of a sequence, and use a separator token $|$ to distinguish the sub-sequence of each single constraint.

\subsubsection{Examples for Typical Tasks}
\label{sec:examples_for_typical_tasks}
In this work, we consider six typical layout generation tasks which has been explored in previous work. We demonstrate how we formulate the constraint sequences for these tasks according to Section~\ref{sec:formulation_principle} as followed.

% Need to be revised to fit the guideline.
\emph{Generation conditioned on types} (Gen-T) is to generate layouts from the element types specified by user constraints.   
We formulate the input constraint sequence by concatenating the single-element constraints of element type, i.e., $S_{\text{Gen-T}}=\{\langle\text{sos}\rangle  c_1 | c_2 | \dots | c_N  \langle\text{eos}\rangle\}$.\footnote{According to the practical experience, we add the special tokens $\langle\text{sos}\rangle$ and $\langle\text{eos}\rangle$ at the start and the end of the sequence, and use a separator token $|$ to distinguish the single constraints.}

\emph{Generation conditioned on types and sizes} (Gen-TS) is to generate layouts when user constraints specify the element types and sizes. We formulate the input sequence by concatenating the type, the width and the height constraints as $S_{\text{Gen-TS}}=\{\langle\text{sos}\rangle  c_1 w_1 h_1 | c_2 w_2 h_2| \dots | c_N w_N h_N  \langle\text{eos}\rangle\}$.

\emph{Generation conditioned on relationships} (Gen-R) generates layouts conditioned on element relationships.
For example, a user could expect to put a large image at the top of a small text box.
% Typically, there are five kinds of position relationships (i.e., above, bottom, left, right, and overlap) and three kinds of size relationships (i.e., smaller, larger and equal)~\cite{LayoutGAN++}.
We formulate a relationship between two elements as $\{c_{k_{2m-1}} k_{2m-1} r_{k_{2m-1}, k_{2m}} c_{k_{2m}} k_{2m}\}$, where $c_{k_{2m-1}}$ and $c_{k_{2m}}$ are the element types, $k_{2m-1}$ and $k_{2m}$ are the indexes for the elements, and $r_{k_{2m-1}, k_{2m}}$ is an extra token introduced to denote one kind of relationships. 
We concatenate the sequences of type constraints and relationship constraints as $S_{\text{Gen-R}}=\{\langle\text{sos}\rangle  c_1 | c_2 | \dots | c_N || c_{k_1} k_1 r_{k_1, k_2} c_{k_2} k_2 | \dots  | c_{k_{2M-1}} k_{2M-1} \\ r_{k_{2M-1}, k_{2M}} c_{k_{2M}} k_{2M} \langle\text{eos}\rangle\}$, where $M$ is the number of relationships.

\emph{Refinement} applies local changes to the elements that need improvements while maintaining the original layout design. 
% The typical improvements include alignment, proper white space, balance etc.
We formulate the input as $S_{\text{Refinement}}=\{\langle\text{sos}\rangle  c_1 x_1 y_1 w_1 h_1 | \dots | c_N x_N y_N w_N h_N  \langle\text{eos}\rangle\}$.

\emph{Completion} aims to complete layout from a set of specified elements.
We formulate the input as $S_{\text{Completion}}=\{\langle\text{sos}\rangle  c_1 x_1 y_1 w_1 h_1 | \dots | c_P x_P y_P w_P h_P  \langle\text{eos}\rangle\}$, where $P$ is the number of known elements.

\emph{Unconstrained generation} (UGen) aims to generate layouts without any user requirements. 
We formulate the input as an empty sequence with necessary special tokens, i.e., $S_{\text{UGen}}=\{\langle\text{sos}\rangle \langle\text{eos}\rangle\}$.

\subsection{Decoding Space Restriction}
\label{sec:decoding_space_restriction}

During the inference, we introduce the decoding space restriction strategy to the decoding process. Algorithm~\ref{alg:decoding_space_restriction} presents the pseudo-code for the decoding space restriction. 

In each decoding step $t$, the decoder predicts the probabilities $P$ of the possible values for current attribute. Then, $P$ will be pruned by two modules. The \emph{ConstraintPruning} prunes the infeasible values which may violate the user given constraints $S$. The \emph{ProbabilityPruning} prunes the values which the probabilities are lower than the threshold $\theta$. 

If all the probabilities in $P'$ are pruned, it means there is no feasible value for current attribute. We propose a backtracking mechanism to solve this problem. When $P'$ is empty and the back time $B[t]$ is less than the max back time $maxBack$, the backtracking mechanism will roll back the decoding process to a previous step $t'$ and restart the prediction. Otherwise, the value $o$ will be sampled from $P'$ and the generation $O$ will be updated. The decoding process will move to the next step by increasing $t$ by 1. 

While the end-of-sequence token EOS is predicted, or the length of $O$ reaches $maxLen$, the generation finishes.
In the following, we will give more details about the two pruning modules and the backtracking mechanism.

\textbf{Constraint Pruning Module.}
The constraint pruning module prunes the value in $P$ which may violate the related constraints.
Take the relationship constraint as example, suppose at step $t$ the decoder predicts possibilities $P$ for the value of the attribute $w_i$ of the $i$-th element, there is a relation constraint $s=\{w_i\leq w_j\}$, which specifies the relative size relationships between element $i$ and $j$. If $w_j$ has not been predicted at step $t$, $w_i$ will not be influenced by $w_j$ in current step and the $P$ will not be pruned. Otherwise, if $w_j$ has been predicted at previous step, the feasible values of $w_i$ will be restricted to $(0,w_j]$ according to $s$, and the infeasible values' probabilities in $P$ will be set as 0. 

% To determine which constraints are related to the current token, since we have defined the format and order for both input and output sequence, we can easily target the constraints in the input sequence to its related attribute in the output sequence. 

\textbf{Probability Pruning Module.}
The probability pruning module checks each value's probability in $P'$. The probabilities that are lower than the predetermined threshold $\theta$ will be pruned by setting as 0. The threshold $\theta$ is tuned to achieve the best performance by task. 

\textbf{Backtracking Mechanism.}
When the probabilities in $P'$ are all set as 0, the backtracking mechanism works to roll back the decoding process to a previous step and restart the prediction. 
The function $Backtracking$ will check why the $P'$ is pruned as empty, and decide which step $t'$ to backtrack to.
One situation is that the current token is restricted by previous token through the constraint. For example, the relation constraint $s=\{w_i\leq w_j\}$ restricts the feasible values of $w_i$ by $w_j$.
In this case, we choose the step of $w_j$ as the backtracking step. 
Otherwise, since all the previous predictions collectively lead to current circumstance, we randomly pick one previous step to back to.

% To avoid the decoder makes same prediction again after backtracking, we set the probability of the value which has been chosen in the last time at step $t'$ before sampling.
% 
\begin{algorithm}[t]
\begin{small}
\caption{Decoding Space Restriction}
\label{alg:decoding_space_restriction}
\KwIn{Encoder hidden state $M$; User constraints $S$.}
\KwOut{Layout sequence $O$.}
\BlankLine

Initialize step index $t$, the back time for each step $B$ and the predicted sequence $O$.

\While{$(O[-1]\neq\textnormal{EOS})$ \textbf{and} $(t$\textless$maxLen)$} {
    $P \leftarrow \textnormal{Decoder}(O, M)[t]$\;
    $P' \leftarrow \textnormal{ConstraintPruning}(P, S)$\;
	$P' \leftarrow \textnormal{ProbabilityPruning}(P', \theta)$\;
	\eIf{$(P'$ \textnormal{is} $\varnothing)$ \textbf{and} $(B[t]$\textless$maxBack)$}
	{
	    $t' \leftarrow \textnormal{Backtracking}(P,S,t)$\;
	    $B[t] \leftarrow B[t] + 1$\;
	    $O \leftarrow O[:t]$\;
	    $t \leftarrow t'$\;
	}
	{
	    $o \leftarrow \textnormal{Sampling}(P')$\;
	    $O \leftarrow O \cup o$\;
	    $t \leftarrow t + 1$\;
	}
}
\end{small}
\end{algorithm}
\begin{table*}[ht]
    \centering
    \renewcommand{\arraystretch}{1.1}
        \resizebox{0.95\textwidth}{!}{
            \begin{tabular}{llcccccccc}
                \specialrule{1.1pt}{0pt}{1pt}
                                            &                   & \multicolumn{4}{c}{RICO}   & \multicolumn{4}{c}{PubLayNet}    \\ 
                                            \cmidrule(l){3-6} \cmidrule(l){7-10}
                Tasks                       & \makecell{Methods}& mIoU $\uparrow$   & FID $\downarrow$  & Align. $\downarrow$   & Overlap $\downarrow$  
                                                                & mIoU $\uparrow$   & FID $\downarrow$  & Align. $\downarrow$   & Overlap $\downarrow$      \\ \hline
                \multirow{4}{*}{Gen-T}      & NDN-none          & 0.35              & 13.76             & 0.56                  & 0.55                                  
                                                                & 0.31              & 35.67             & 0.35                  & 0.17                      \\ 
                                            & LayoutGAN++       & 0.298             & 5.954             & 0.261                 & 0.620                 
                                                                & 0.297             & 14.875            & 0.124                 & 0.148                     \\ 
                                            & BLT               & 0.216             & 25.633            & 0.150                 & 0.983                               
                                                                & 0.140             & 38.684            & 0.036                 & 0.196                     \\  
                                            & LayoutFormer++    & \textbf{0.432}    & \textbf{1.096}    & 0.230         & \textbf{0.530}                      
                                                                & \textbf{0.348}    & \textbf{8.411}    & \textbf{0.020}        & \textbf{0.008}            \\ % 
                                            \hline

                \multirow{2}{*}{Gen-TS}     & BLT               & 0.604             & 0.951             & 0.181                 & 0.660                          
                                                                & 0.428             & 7.914             & 0.021                 & 0.419                                   \\ 
                                            & LayoutFormer++    & \textbf{0.620}    & \textbf{0.757}    & 0.202                 & \textbf{0.542}           
                                                                & \textbf{0.471}    & \textbf{0.720}    & 0.024                 & \textbf{0.037}            \\ % 
                                            \hline
                \multirow{3}{*}{Gen-R}      & NDN               & 0.36              & -                 & 0.56                  & -                     
                                                                & 0.31              & -                 & 0.36                  & -                                      \\ 
                                            & CLG-LO            & 0.286             & 8.898             & 0.311                 & 0.615                 
                                                                & 0.277             & 19.738            & 0.123                 & 0.200                       \\  
                                            & LayoutFormer++    & \textbf{0.424}    & \textbf{5.972}    & 0.332                 & \textbf{0.537}                     
                                                                & \textbf{0.353}    & \textbf{4.954}    & \textbf{0.025}        & \textbf{0.076}           \\ % 
                                            \hline
                \multirow{2}{*}{Refinement} & RUITE             & 0.811             & 0.107             & 0.133                 & 0.483                  
                                                                & 0.781             & 0.061             & 0.029                 & 0.020.                 \\
                                            & LayoutFormer++    & \textbf{0.816}    & \textbf{0.032}    & \textbf{0.123}        & 0.489                 
                                                                & \textbf{0.785}    & 0.086             & \textbf{0.024}        & \textbf{0.006}        \\
                                            \hline
                \multirow{2}{*}{Completion} & LayoutTransformer & 0.363             & 6.679             & 0.194                 & 0.478                 
                                                                & 0.077             & 14.769            & 0.019                 & 0.0013                \\
                                            & LayoutFormer++    & \textbf{0.732}    & \textbf{4.574}    & \textbf{0.077}        & 0.487                 
                                                                & \textbf{0.471}    & \textbf{10.251}   & 0.020                 & 0.0022                \\
                                            \hline
                \multirow{4}{*}{UGen}       & LayoutTransformer & 0.439             & 22.884            & 0.052                 & 0.471                  
                                                                & 0.062             & 36.304            & 0.031                 & 0.0009                \\
                                            & VTN               & 0.686             & 76.064            & 0.461                 & 0.694                  
                                                                & 0.210             & 103.373           & 0.205                 & 0.211                \\ 
                                            & Coarse2Fine       & 0.360             & 46.483            & 0.128                 & 0.676                  
                                                                & 0.361             & 50.854            & 0.221                 & 0.142                  \\ 
                                            & LayoutFormer++    & \textbf{0.742}    & \textbf{19.688}   & \textbf{0.047}        & 0.547                  
                                                                & \textbf{0.417}    & 46.522            & \textbf{0.029}        & \textbf{0.0009}        \\
                \specialrule{1.1pt}{1pt}{0pt}
            \end{tabular}}
    \setlength{\belowcaptionskip}{-0.35cm}
    \caption{Quantitative comparisons with existing approaches on six layout generation tasks.}
    \label{tab:flexibility}
\end{table*}

\begin{figure*}[htb]
    \centering
      \includegraphics[width=0.96\linewidth]{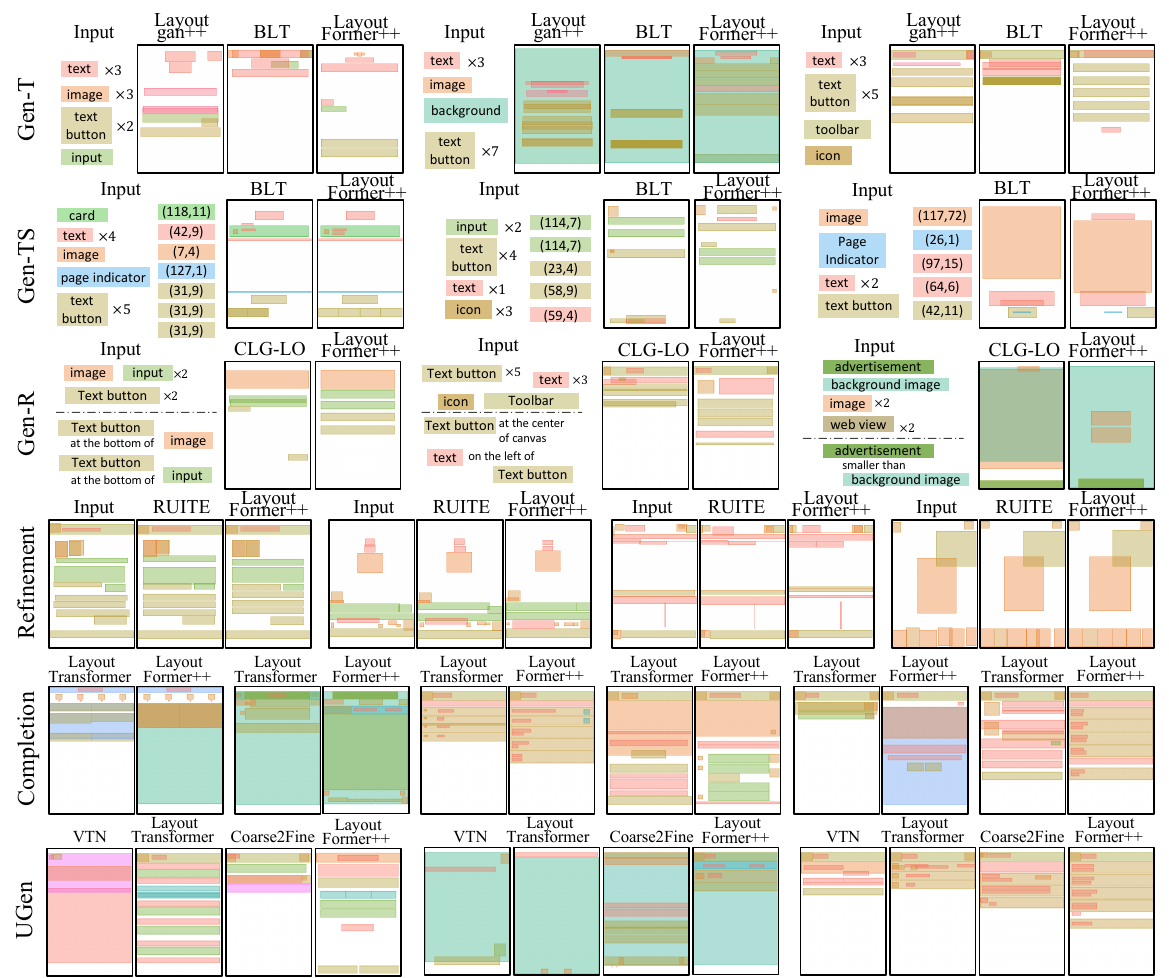}
    \caption{Qualitative results on RICO.}
    \label{fig:qualitative_rico}
\end{figure*}
  
\begin{figure*}[htb]
    \centering
      \includegraphics[width=0.96\linewidth]{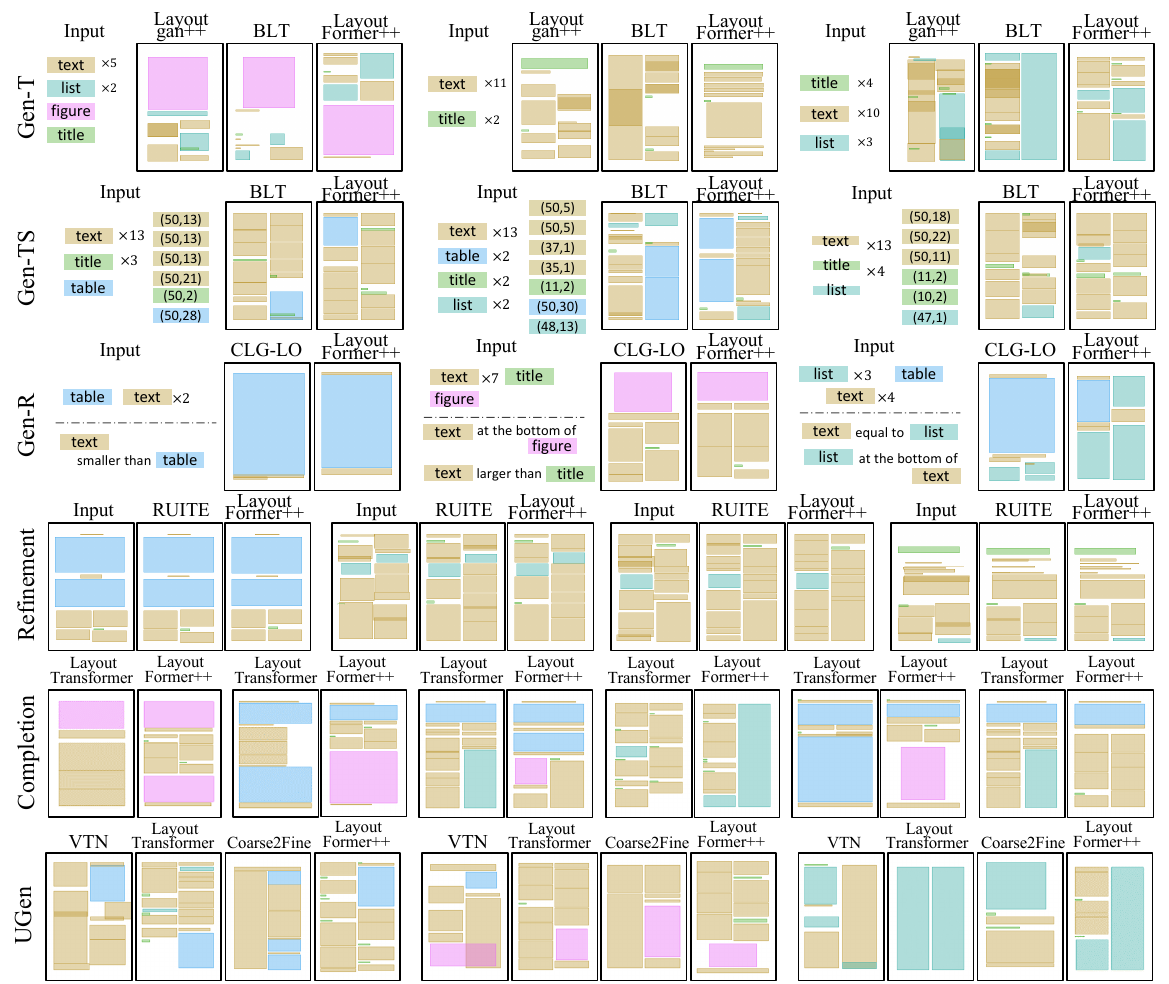}
    \caption{Qualitative results on PubLayNet.}
    \label{fig:qualitative_pln}
\end{figure*}

\section{Experiments}
\label{sec:experiments}

\subsection{Setups}
\label{sec:setups}

\textbf{Datasets.}
We compare LayoutFormer++ to existing approaches on two public datasets.
\emph{RICO}~\cite{rico} is a dataset of mobile app UI that contains $66$K+ UI layouts with $25$ element types. 
\emph{PubLayNet}~\cite{publaynet} contains $360$K+ document layouts with $5$ element types. 
On both datasets, there are a few layouts with quite a lot of elements, which easily leads to an out-of-memory problem. 
Existing studies use multifarious rules to filter out these layouts.
In our experiments, we simply filter out the layouts with more than $20$ elements.
For RICO, we use $90\%$, $5\%$ and $5\%$ of data for training, validation and testing. For PubLayNet, we use $95\%$ and $5\%$ of official training split for training and validation, and the official validation split for testing.

\textbf{Baselines.}
We try our best to reproduce all the existing approaches.
Even so, we regret to not include some approaches for the following reasons.
First, due to the missing implementation details and hyperparameter settings, we fail to reproduce some approaches~\cite{LayoutGAN,attributelayoutgan,READ,layoutmcl}.
Second, a few approaches consider data-specific attributes and are difficult to be applied to arbitrary datasets about graphic layouts~\cite{content-aware,CanvasVAE}.
Ultimately, we compare LayoutFormer++ against 1) \emph{NDN-none}~\cite{NDN}, \emph{LayoutGAN++}~\cite{LayoutGAN++} and \emph{BLT}~\cite{BLT} on Gen-T, 2) \emph{BLT}~\cite{BLT} on Gen-TS, 3) \emph{NDN}~\cite{NDN} and \emph{CLG-LO}~\cite{LayoutGAN++} on Gen-R, 4) \emph{RUITE}~\cite{RUITE} on refinement, 5) \emph{LayoutTransformer}~\cite{LayoutTransformer} on completion, and 6) \emph{VTN}~\cite{VTN}, \emph{Coarse2Fine}~\cite{coarse2fine} and \emph{LayoutTransformer}~\cite{LayoutTransformer} on UGen.

\textbf{Evaluation Metrics.}
We adopt the metrics proposed by existing works for comprehensive evaluation.
 
\emph{Maximum Intersection over Union} (mIoU) measures the similarity between the generated layouts and the real layouts, which is based on the averaged IoU of bounding boxes. 
We use the same implementation as \cite{LayoutGAN++}.

\emph{Alignment} (Align.) measures whether the elements are well-aligned. 
We modify the metric from \cite{attributelayoutgan} by normalizing it by the number of elements.

\emph{Overlap} measures the abnormal overlap area between elements. 
We improve the metric from \cite{attributelayoutgan} by ignoring the elements that serve as a background or padding, e.g., card, background and modal on RICO. 

\emph{Frechet Inception Distance} (FID) describes the distribution difference between real and generated layouts. Following \cite{LayoutGAN++}, we train a neural network to convert the layouts into representative features and then calculate the distribution difference based on learned features.

\emph{Constraint Violation Rate} (Vio.\%) measures the rate of the violated constraints. We follow the implementation of \cite{LayoutGAN++} for Gen-R. For other tasks, we calculate Vio.\% by dividing the number of the violated constraints by the total number of constraints for all the evaluated layouts.

Among the above metrics, mIoU, Align., Overlap and FID evaluate the generation quality, and Vio.\% evaluate the constraint satisfaction. A larger value for mIoU indicates better performance, while smaller values for other metrics indicate better performance.

\textbf{Implementation Details.}
We implement LayoutFormer++ by PyTorch~\cite{PyTorch}.
The model is trained using the Adam optimizer~\cite{Adam} with NVIDIA V100 GPUs. 
For Transformer blocks, we use $8$ layers, $8$ heads for multi-attention, $512$ embedding dimensions and $2048$ feed-forward dimensions.
Other hyper-parameters, e.g., the batch size, the learning rate, and the threshold $\theta$ in decoding space restriction, are tuned to achieve the best performance on the validation set\footnote{Details of the hyper-parameters can be find in supplemental material.}.
For Gen-R, following CLG-LO~\cite{LayoutGAN++}, we randomly sample $10\%$ element relationships as the input.
For refinement, following RUITE~\cite{RUITE}, we synthesize the input by adding random noise to the position and size of each element, which is sampled from a normal distribution with the mean $0$ and the standard variance $0.01$.
% At the inference time, on refinement, we simply use greedy sampling, and on the other subtasks, we use top-k sampling with $k = 10$ and temperature $\tau = 0.5$ to generate multiple layouts.

\subsection{Evaluations on Sufficient Flexibility}
\label{sec:evaluations_on_sufficient_flexibility}

To demonstrate the sufficient flexibility of LayoutFormer++, we compare LayoutFormer++ with existing approaches on all six layout generation tasks.

Table~\ref{tab:flexibility} shows the quantitative comparisons to existing works\footnote{As we fail to reproduce NDN by ourselves, we directly use the results from \cite{LayoutGAN++} in Table~\ref{tab:flexibility}. It is evaluated on simple datasets that only consider layouts with less than $10$ elements.}. 
It should be noted that the existing approaches can only handle up to two tasks, while LayoutFormer++ can handle all the six tasks.
We highlight the results of LayoutFormer++ by bold when it achieves the best performance among all the existing approaches in Table~\ref{tab:flexibility}. According to the results, LayoutFormer++ achieves significantly better performance than the baselines on most metrics. This demonstrates that LayoutFormer++ not only can flexibly handle diverse user constraints, but also has advantages in the generation ability over existing approaches.

Figure~\ref{fig:qualitative_rico} and \ref{fig:qualitative_pln} show qualitative comparisons on RICO and PubLayNet.
% \footnote{More qualitative results of LayoutFormer++ are available in the supplemental material.}. 
% For Gen-T, Gen-TS and Gen-R, in each group, the first column shows constraints from the user. 
% For refinement, in each group, the first column shows a layout draft from users. 
% For completion, layouts in the same group are generated from the same element. 
Compared to baselines, LayoutFormer++ generates layouts with better spacing, less misalignment, and fewer unreasonable overlaps.
\subsection{Evaluations on Good Controllability}
\label{sec:evaluations_on_good_controllability}

\begin{table*}[tb]
    \centering
    \renewcommand{\arraystretch}{1.1}
    \resizebox{0.93\textwidth}{!}{
    \begin{tabular}{lL{1.2cm}L{2.1cm}cccccccccc}
        \specialrule{1.1pt}{0pt}{1pt}
                                &                   & \multicolumn{5}{c}{RICO}   & \multicolumn{5}{c}{PubLayNet}    \\ 
                                \cmidrule(l){4-8} \cmidrule(l){9-13}
        Tasks                   & \multicolumn{2}{c}{Method}    & mIoU $\uparrow$   & FID $\downarrow$  & Align. $\downarrow$   & Overlap $\downarrow$  & Vio. \% $\downarrow$ 
                                                                & mIoU $\uparrow$   & FID $\downarrow$  & Align. $\downarrow$   & Overlap $\downarrow$  & Vio. \% $\downarrow$  \\ \hline
        \multirow{4}{*}{Gen-T}  
        &\multicolumn{2}{c}{LayoutGAN++}                        & 0.298             & 5.954             & 0.261                 & 0.620                 & 0.
                                                                & 0.297             & 14.875            & 0.124                 & 0.148                 & 0.    \\ \cmidrule{2-13} 
        & \multirow{3}{*}{\makecell[r]{Layout\\Former++}}            
                                & \ \ Full                      & \textbf{0.432}        & \textbf{1.096} 
                                                                & \textbf{0.230}        & \textbf{0.530}     & \textbf{0.}                
                                                                & \textbf{0.348}        & \textbf{8.411} 
                                                                & \textbf{0.020}        & \textbf{0.008}     & \textbf{0.}  \\ % 
        &                       & - Back                        & 0.431             & 1.320             & 0.272                 & 0.550                 & 0.                  
                                                                & 0.345             & 9.367             & 0.020                 & 0.009                 & 0.                  \\ % 
        &                       & - Back\&Prune                 & 0.439             & 1.392             & 0.206                 & 0.545                 & 5.5          
                                                                & 0.345             & 9.373             & 0.020                 & 0.009                 & 0.05                  \\ % 
        \hline

        \multirow{4}{*}{Gen-TS} 
        & \multicolumn{2}{c}{BLT}                               & 0.604             & 0.951             & 0.181                 & 0.660                 & 0.         
                                                                & 0.428             & 7.914             & 0.021                 & 0.419                 & 0.    \\ \cmidrule{2-13}
        & \multirow{3}{*}{\makecell[r]{Layout\\Former++}}
                                & \ \ Full                      & \textbf{0.620}        & \textbf{0.757} 
                                                                & 0.202                 & \textbf{0.542}     & \textbf{0.}
                                                                & \textbf{0.471}        & \textbf{0.720}    
                                                                & 0.024                 & \textbf{0.037}     & \textbf{0.}         \\ % 
        &                       & - Back                        & 0.613             & 0.782             & 0.206                 & 0.543                 & 0.                  
                                                                & 0.464             & 0.903             & 0.026                 & 0.044                 & 0.                  \\ % 
        &                       & - Back\&Prune                 & 0.613             & 0.801             & 0.206                 & 0.545                 & $\approx$0.          
                                                                & 0.464             & 0.903             & 0.026                 & 0.044                 & $\approx$0.                 \\ % 
        \hline
        \multirow{4}{*}{Gen-R} 
        & \multicolumn{2}{c}{CLG-LO}                            & 0.286             & 8.898             & 0.311                 & 0.615                 & 3.66
                                                                & 0.277             & 19.738            & 0.123                 & 0.200                 & 6.66  \\ \cmidrule{2-13}
        & \multirow{3}{*}{\makecell[r]{Layout\\Former++}} 
                                & \ \ Full                      & \textbf{0.424}        & \textbf{5.972} 
                                                                & 0.332                             & \textbf{0.537}     & 11.84                
                                                                & \textbf{0.353}        & \textbf{4.954}    
                                                                & \textbf{0.025}                    & \textbf{0.076}                 & \textbf{3.9}    \\ % 
        &                       & - Back                        & 0.419             & 8.604             & 0.284                 & 0.544                 & 12.75                  
                                                                & 0.352             & 5.152             & 0.023                 & 0.075                 & 5.70                 \\ % 
        &                       & - Back\&Prune                 & 0.458             & 5.126             & 0.221                 & 0.546                 & 33.04          
                                                                & 0.358             & 4.620             & 0.022                 & 0.030                 & 16.09                  \\ % 
        
        \specialrule{1.1pt}{1pt}{0pt}
    \end{tabular}}
    \caption{Comparisons with baselines and the model variants to evaluate the controllability.}
    \label{tab:controllability}
\end{table*}
In this section, we compare LayoutFormer++ with LayoutGAN++~\cite{LayoutGAN}, BLT~\cite{BLT} and CLG-LO~\cite{LayoutGAN++}, which pay attention to the constraint satisfaction. These approaches handle Gen-T, Gen-TS and Gen-R respectively.

% For each task in Table~\ref{tab:controllability}, the first and second lines show the comparison between the baselines with LayoutFormer++ on both the quality metrics and the Vio.\%. 
We bold the results of LayoutFormer++ in Table~\ref{tab:controllability} when it outperforms the baselines. For Gen-T and Gen-TS, since all approaches have no constraint violation, we focus on the performance of quality. As the results, LayoutFormer++ outperforms the baselines on both two datasets. 
For Gen-R, LayoutFormer++ achieves better performance on both quality metrics and Vio.\%, except a worse Vio.\% than CLG-LO on RICO. However, CLG-LO does not make a good trade-off: the quality of layouts generated by CLG-LO is significantly worse than that of LayoutFormer++.
This demonstrates that LayoutFormer++ achieves the best controllability among existing approaches.

% Besides, we develop the ablation studies to show the effectiveness of the decoding space restriction strategy in achieving good controllability. 
Besides, we compare LayoutFormer++ framework with two variants. The first denoted as \emph{-Back} indicates LayoutFormer++ without backtracking mechanism. The second denoted as \emph{-Back\&Prune.} is LayoutFormer++ without both pruning modules and the backtracking mechanism. For each task in Table~\ref{tab:controllability}, the last three lines show the comparison between the variants. 
Comparing \emph{-Full} with \emph{-Back}, we find that after disabling the backtracking mechanism, the generation quality decreases. And on Gen-R, the Vio.\% also gets worse. 
Comparing \emph{-Full} with \emph{-Back\&Prune}, we find that the Vio.\% significantly decreases. 
% We underline the quality results of \emph{-Full} when they are better than any of the two variants, and double underline the Vio.\% when they achieve the best.
It shows that the full model consistently decreases the violation while having little impact on quality, which demonstrates that both the pruning modules and the backtracking mechanism play important roles in helping LayoutFormer++ achieve good controllability.
\section{Conclusion}
\label{sec:conclusion}
In this work, we propose \emph{LayoutFormer++} for conditional layout generation.
To achieve the sufficient flexibility, we propose constraint serialization to represent different user constraints by the same sequence format. 
To achieve the good controllability, we propose decoding space restriction to prune the predicted distribution by disabling the options that violate the constraints or lead to bad quality. 
Experiments show that LayoutFormer++ can flexibly handle different layout generation tasks with better generation quality and less constraint violation compared to the existing task-specific approaches. 
In the future, we plan to investigate more practical user requirements for layout design. 
Besides, as LayoutFormer++ enables a unified way to handle different layout generation tasks, it is possible to develop a powerful pretrained model for layout generation in the future.
% which we leave as an important future work.

%%%%%%%%% REFERENCES
{\small
\bibliographystyle{ieee_fullname}
\bibliography{egbib}

\begin{thebibliography}{1}\itemsep=-1pt

\bibitem{aribandi2022ext}
Vamsi Aribandi, Yi Tay, Tal Schuster, Jinfeng Rao, Huaixiu~Steven Zheng,
  Sanket~Vaibhav Mehta, Honglei Zhuang, Vinh~Q. Tran, Dara Bahri, Jianmo Ni,
  Jai Gupta, Kai Hui, Sebastian Ruder, and Donald Metzler.
\newblock Ext5: Towards extreme multi-task scaling for transfer learning.
\newblock In {\em International Conference on Learning Representations}, 2022.

\bibitem{t5}
Colin Raffel, Noam Shazeer, Adam Roberts, Katherine Lee, Sharan Narang, Michael
  Matena, Yanqi Zhou, Wei Li, and Peter~J. Liu.
\newblock Exploring the limits of transfer learning with a unified text-to-text
  transformer.
\newblock {\em Journal of Machine Learning Research}, 21(140):1--67, 2020.

\bibitem{sanh2022multitask}
Victor Sanh, Albert Webson, Colin Raffel, Stephen Bach, Lintang Sutawika, Zaid
  Alyafeai, Antoine Chaffin, Arnaud Stiegler, Arun Raja, Manan Dey, M~Saiful
  Bari, Canwen Xu, Urmish Thakker, Shanya~Sharma Sharma, Eliza Szczechla,
  Taewoon Kim, Gunjan Chhablani, Nihal Nayak, Debajyoti Datta, Jonathan Chang,
  Mike Tian-Jian Jiang, Han Wang, Matteo Manica, Sheng Shen, Zheng~Xin Yong,
  Harshit Pandey, Rachel Bawden, Thomas Wang, Trishala Neeraj, Jos Rozen,
  Abheesht Sharma, Andrea Santilli, Thibault Fevry, Jason~Alan Fries, Ryan
  Teehan, Teven~Le Scao, Stella Biderman, Leo Gao, Thomas Wolf, and Alexander~M
  Rush.
\newblock Multitask prompted training enables zero-shot task generalization.
\newblock In {\em International Conference on Learning Representations}, 2022.

\end{thebibliography}


\begin{thebibliography}{10}\itemsep=-1pt

\bibitem{anderson-etal-2017-guided}
Peter Anderson, Basura Fernando, Mark Johnson, and Stephen Gould.
\newblock Guided open vocabulary image captioning with constrained beam search.
\newblock In {\em Proceedings of the 2017 Conference on Empirical Methods in
  Natural Language Processing}, pages 936--945, Copenhagen, Denmark, Sept.
  2017. Association for Computational Linguistics.

\bibitem{VTN}
Diego~Martin Arroyo, Janis Postels, and Federico Tombari.
\newblock Variational transformer networks for layout generation.
\newblock In {\em Proceedings of the IEEE/CVF Conference on Computer Vision and
  Pattern Recognition (CVPR)}, pages 13642--13652, June 2021.

\bibitem{LayoutTransformer}
Kamal Gupta, Justin Lazarow, Alessandro Achille, Larry~S. Davis, Vijay
  Mahadevan, and Abhinav Shrivastava.
\newblock Layouttransformer: Layout generation and completion with
  self-attention.
\newblock In {\em Proceedings of the IEEE/CVF International Conference on
  Computer Vision (ICCV)}, pages 1004--1014, October 2021.

\bibitem{hokamp-liu-2017-lexically}
Chris Hokamp and Qun Liu.
\newblock Lexically constrained decoding for sequence generation using grid
  beam search.
\newblock In {\em Proceedings of the 55th Annual Meeting of the Association for
  Computational Linguistics (Volume 1: Long Papers)}, pages 1535--1546,
  Vancouver, Canada, July 2017. Association for Computational Linguistics.

\bibitem{Holtzman2020The}
Ari Holtzman, Jan Buys, Li Du, Maxwell Forbes, and Yejin Choi.
\newblock The curious case of neural text degeneration.
\newblock In {\em International Conference on Learning Representations}, 2020.

\bibitem{hu-etal-2019-improved}
J.~Edward Hu, Huda Khayrallah, Ryan Culkin, Patrick Xia, Tongfei Chen, Matt
  Post, and Benjamin Van~Durme.
\newblock Improved lexically constrained decoding for translation and
  monolingual rewriting.
\newblock In {\em Proceedings of the 2019 Conference of the North {A}merican
  Chapter of the Association for Computational Linguistics: Human Language
  Technologies, Volume 1 (Long and Short Papers)}, pages 839--850, Minneapolis,
  Minnesota, June 2019. Association for Computational Linguistics.

\bibitem{coarse2fine}
Zhaoyun Jiang, Shizhao Sun, Jihua Zhu, Jian-Guang Lou, and Dongmei Zhang.
\newblock Coarse-to-fine generative modeling for graphic layouts.
\newblock In {\em AAAI'22}, February 2022.

\bibitem{LayoutGAN++}
Kotaro Kikuchi, Edgar Simo-Serra, Mayu Otani, and Kota Yamaguchi.
\newblock Constrained graphic layout generation via latent optimization.
\newblock In {\em Proceedings of the 29th ACM International Conference on
  Multimedia}, MM '21, page 88–96, New York, NY, USA, 2021. Association for
  Computing Machinery.

\bibitem{Adam}
Diederik~P. Kingma and Jimmy Ba.
\newblock Adam: A method for stochastic optimization.
\newblock In {\em ICLR}, 2015.

\bibitem{BLT}
Xiang Kong, Lu Jiang, Huiwen Chang, Han Zhang, Yuan Hao, Haifeng Gong, and
  Irfan Essa.
\newblock Blt: Bidirectional layout transformer for controllable layout
  generation, 2021.

\bibitem{krishnamurthy-etal-2017-neural}
Jayant Krishnamurthy, Pradeep Dasigi, and Matt Gardner.
\newblock Neural semantic parsing with type constraints for semi-structured
  tables.
\newblock In {\em Proceedings of the 2017 Conference on Empirical Methods in
  Natural Language Processing}, pages 1516--1526, Copenhagen, Denmark, Sept.
  2017. Association for Computational Linguistics.

\bibitem{NDN}
Hsin-Ying Lee, Lu Jiang, Irfan Essa, Phuong~B. Le, Haifeng Gong, Ming-Hsuan
  Yang, and Weilong Yang.
\newblock Neural design network: Graphic layout generation with constraints.
\newblock In Andrea Vedaldi, Horst Bischof, Thomas Brox, and Jan-Michael Frahm,
  editors, {\em Computer Vision -- ECCV 2020}, pages 491--506, Cham, 2020.
  Springer International Publishing.

\bibitem{LayoutGAN}
Jianan Li, Jimei Yang, Aaron Hertzmann, Jianming Zhang, and Tingfa Xu.
\newblock Layoutgan: Generating graphic layouts with wireframe discriminators,
  2019.

\bibitem{attributelayoutgan}
Jianan Li, Jimei Yang, Jianming Zhang, Chang Liu, Christina Wang, and Tingfa
  Xu.
\newblock Attribute-conditioned layout gan for automatic graphic design.
\newblock {\em IEEE Transactions on Visualization and Computer Graphics},
  27(10):4039–4048, aug 2021.

\bibitem{rico}
Thomas~F. Liu, Mark Craft, Jason Situ, Ersin Yumer, Radomir Mech, and Ranjitha
  Kumar.
\newblock Learning design semantics for mobile apps.
\newblock In {\em The 31st Annual ACM Symposium on User Interface Software and
  Technology}, UIST '18, pages 569--579, New York, NY, USA, 2018. ACM.

\bibitem{lu-etal-2022-neurologic}
Ximing Lu, Sean Welleck, Peter West, Liwei Jiang, Jungo Kasai, Daniel Khashabi,
  Ronan Le~Bras, Lianhui Qin, Youngjae Yu, Rowan Zellers, Noah~A. Smith, and
  Yejin Choi.
\newblock {N}euro{L}ogic a*esque decoding: Constrained text generation with
  lookahead heuristics.
\newblock In {\em Proceedings of the 2022 Conference of the North American
  Chapter of the Association for Computational Linguistics: Human Language
  Technologies}, pages 780--799, Seattle, United States, July 2022. Association
  for Computational Linguistics.

\bibitem{lu-etal-2021-neurologic}
Ximing Lu, Peter West, Rowan Zellers, Ronan Le~Bras, Chandra Bhagavatula, and
  Yejin Choi.
\newblock {N}euro{L}ogic decoding: (un)supervised neural text generation with
  predicate logic constraints.
\newblock In {\em Proceedings of the 2021 Conference of the North American
  Chapter of the Association for Computational Linguistics: Human Language
  Technologies}, pages 4288--4299, Online, June 2021. Association for
  Computational Linguistics.

\bibitem{layoutmcl}
David~D. Nguyen, Surya Nepal, and Salil~S. Kanhere.
\newblock Diverse multimedia layout generation with multi choice learning.
\newblock In {\em Proceedings of the 29th ACM International Conference on
  Multimedia}, MM '21, page 218–226, New York, NY, USA, 2021. Association for
  Computing Machinery.

\bibitem{PyTorch}
Adam Paszke, Sam Gross, Francisco Massa, Adam Lerer, James Bradbury, Gregory
  Chanan, Trevor Killeen, Zeming Lin, Natalia Gimelshein, Luca Antiga, Alban
  Desmaison, Andreas Kopf, Edward Yang, Zachary DeVito, Martin Raison, Alykhan
  Tejani, Sasank Chilamkurthy, Benoit Steiner, Lu Fang, Junjie Bai, and Soumith
  Chintala.
\newblock Pytorch: An imperative style, high-performance deep learning library.
\newblock In H. Wallach, H. Larochelle, A. Beygelzimer, F. d\textquotesingle
  Alch\'{e}-Buc, E. Fox, and R. Garnett, editors, {\em Advances in Neural
  Information Processing Systems 32}, pages 8024--8035. Curran Associates,
  Inc., 2019.

\bibitem{READ}
Akshay~Gadi Patil, Omri Ben-Eliezer, Or Perel, and Hadar Averbuch-Elor.
\newblock Read: Recursive autoencoders for document layout generation.
\newblock In {\em Proceedings of the IEEE/CVF Conference on Computer Vision and
  Pattern Recognition (CVPR) Workshops}, June 2020.

\bibitem{post-vilar-2018-fast}
Matt Post and David Vilar.
\newblock Fast lexically constrained decoding with dynamic beam allocation for
  neural machine translation.
\newblock In {\em Proceedings of the 2018 Conference of the North {A}merican
  Chapter of the Association for Computational Linguistics: Human Language
  Technologies, Volume 1 (Long Papers)}, pages 1314--1324, New Orleans,
  Louisiana, June 2018. Association for Computational Linguistics.

\bibitem{qin2022cold}
Lianhui Qin, Sean Welleck, Daniel Khashabi, and Yejin Choi.
\newblock {COLD} decoding: Energy-based constrained text generation with
  langevin dynamics.
\newblock In Alice~H. Oh, Alekh Agarwal, Danielle Belgrave, and Kyunghyun Cho,
  editors, {\em Advances in Neural Information Processing Systems}, 2022.

\bibitem{RUITE}
Soliha Rahman, Vinoth~Pandian Sermuga~Pandian, and Matthias Jarke.
\newblock Ruite: Refining ui layout aesthetics using transformer encoder.
\newblock In {\em 26th International Conference on Intelligent User Interfaces
  - Companion}, IUI '21 Companion, page 81–83, New York, NY, USA, 2021.
  Association for Computing Machinery.

\bibitem{scholak-etal-2021-picard}
Torsten Scholak, Nathan Schucher, and Dzmitry Bahdanau.
\newblock {PICARD}: Parsing incrementally for constrained auto-regressive
  decoding from language models.
\newblock In {\em Proceedings of the 2021 Conference on Empirical Methods in
  Natural Language Processing}, pages 9895--9901, Online and Punta Cana,
  Dominican Republic, Nov. 2021. Association for Computational Linguistics.

\bibitem{shin-van-durme-2022-shot}
Richard Shin and Benjamin Van~Durme.
\newblock Few-shot semantic parsing with language models trained on code.
\newblock In {\em Proceedings of the 2022 Conference of the North American
  Chapter of the Association for Computational Linguistics: Human Language
  Technologies}, pages 5417--5425, Seattle, United States, July 2022.
  Association for Computational Linguistics.

\bibitem{Transformer}
Ashish Vaswani, Noam Shazeer, Niki Parmar, Jakob Uszkoreit, Llion Jones,
  Aidan~N Gomez, \L~ukasz Kaiser, and Illia Polosukhin.
\newblock Attention is all you need.
\newblock In I. Guyon, U.~Von Luxburg, S. Bengio, H. Wallach, R. Fergus, S.
  Vishwanathan, and R. Garnett, editors, {\em Advances in Neural Information
  Processing Systems}, volume~30. Curran Associates, Inc., 2017.

\bibitem{CanvasVAE}
Kota Yamaguchi.
\newblock Canvasvae: Learning to generate vector graphic documents.
\newblock In {\em Proceedings of the IEEE/CVF International Conference on
  Computer Vision (ICCV)}, pages 5481--5489, October 2021.

\bibitem{zhang-etal-2020-language-generation}
Maosen Zhang, Nan Jiang, Lei Li, and Yexiang Xue.
\newblock Language generation via combinatorial constraint satisfaction: A tree
  search enhanced {M}onte-{C}arlo approach.
\newblock In {\em Findings of the Association for Computational Linguistics:
  EMNLP 2020}, pages 1286--1298, Online, Nov. 2020. Association for
  Computational Linguistics.

\bibitem{content-aware}
Xinru Zheng, Xiaotian Qiao, Ying Cao, and Rynson W.~H. Lau.
\newblock Content-aware generative modeling of graphic design layouts.
\newblock {\em ACM Trans. Graph.}, 38(4), jul 2019.

\bibitem{publaynet}
Xu Zhong, Jianbin Tang, and Antonio~Jimeno Yepes.
\newblock Publaynet: largest dataset ever for document layout analysis.
\newblock {\em arXiv preprint arXiv:1908.07836}, 2019.

\end{thebibliography}
}

\end{document}

% \title{\LaTeX\ Author Guidelines for \confName~Proceedings}
% \title{Multi-Constrained Layout Generation with Layout Constrained Decoding}
\title{Supplementary Material for \\ ``LayoutFormer++: Conditional Graphic Layout Generation via Constraint Serialization and Decoding Space Restriction''}
% \and
\maketitle

\section{Implementation Details}
\label{sec:hyper_params}

\subsection{Training}

The training epochs and batch sizes of LayoutFormer++ for each layout generation task are shown in Table~\ref{Tab:params}.
On both RICO and PubLayNet, we use optimizer Adafactor with a learning rate of 0.0001. 
We use the learning rate warmup. The numbers of the warmup steps for each model are also shown in Table~\ref{Tab:params}.

\begin{table}[!htp]
\centering
\renewcommand{\arraystretch}{1.2}
\begin{small}
    \begin{tabular}{lcccccc}
        \specialrule{1.1pt}{0pt}{1pt}
                    & \multicolumn{3}{c}{RICO}              & \multicolumn{3}{c}{PubLayNet}         \\ \cmidrule(l){2-4} \cmidrule(l){5-7}
        Task	    & Epoch	& Batch Size	& Warmup Steps	& Epoch	& Batch Size    & Warmup Steps  \\  \specialrule{0.8pt}{1pt}{1pt}
        Gen-T       & 100   & 32            & 1000          & 60    & 100           & 1000          \\
        Gen-TS      & 100   & 32            & 1000          & 60    & 100           & 4000          \\
        Gen-R       & 150   & 16            & 1000          & 60    & 64            & 3000          \\
        Completion  & 100   & 32            & 1000          & 60    & 100           & 3000          \\
        Refinement  & 100   & 32            & 1000          & 60    & 100           & 2000          \\
        UGen        & 100   & 32            & 1000          & 60    & 100           & 1000          \\
        \specialrule{1.1pt}{1pt}{0pt}
    \end{tabular}
\end{small}
\caption{Hyper-parameters for training LayoutFormer++.}
\label{Tab:params}
\end{table}

\subsection{Inference}

During the inference stage, we leverage top-k sampling with $k = 10$ and temperature $\tau = 0.7$ for LayoutFormer++ to generate diverse layouts for all tasks, except refinement which simply uses greedy sampling. 
For the decoding space restriction strategy, to avoid the decoder make the same prediction after backtracking, we change the top-k temperature $\tau$ from 0.7 to 1.5 to smooth the sampling distribution at the step that the decoding process backs to.
The max back time $maxBack$ of the backtracking mechanism, and the threshold $\theta$ of the probability pruning module, are tuned to achieve the best performance for each task (i.e., Gen-T, Gen-TS, and Gen-R) and dataset, which are shown in Table~\ref{Tab:params_in_decoding}.

\begin{table}[!htp]
\centering
\renewcommand{\arraystretch}{1.2}
\begin{small}
    \begin{tabular}{ccccccc}
        \specialrule{1.1pt}{0pt}{1pt}
                      & \multicolumn{3}{c}{RICO}                  & \multicolumn{3}{c}{PubLayNet}         \\ \cmidrule(l){2-4} \cmidrule(l){5-7}
          Tasks       & Gen-T     & Gen-TS    & Gen-R     & Gen-T     & Gen-TS    & Gen-R \\
                      \specialrule{0.8pt}{1pt}{1pt}
          $maxBack$   & 5         & 5         & 5         & 5         & 3         & 3     \\
          $\theta$    & 0.8       & 0.3       & 0.2       & 0.3       & 0.3       & 0.2   \\
        \specialrule{1.1pt}{2pt}{0pt}
    \end{tabular}
\end{small}
\caption{Hyper-parameters of the Decoding Space Restriction Strategy during the  Inference}
\label{Tab:params_in_decoding}
\end{table}

\newpage

\section{More Discussion on Constraint Serialization}
\label{sec:sufficient_flecibility}

\subsection{All-in-one Layout Generation Model}
\label{sec:mixed_tasks}

The unification of the constraint format and the model architecture opens up a chance for training a single model to serve all tasks simultaneously. This is of great benefit. It saves a lot of deployment efforts in practice since there is a single set of model weights for flexibly serving multiple tasks. 

To achieve this goal, we train LayoutFormer++ for the mixed-task generation, denoted as LayoutFormer++(M). Following \cite{t5,aribandi2022ext,sanh2022multitask}, we prepend a task indicator token to each input sequence and combine the training data of all tasks with temperature mixing. The data sample weights of each task for constructing a mixed dataset are shown in Table~\ref{Tab:sample_weight}, which are the same on both Rico and PubLayNet. In practice, we find that slightly increasing the weights of Refinement and Gen-TS can better balance the performance for all tasks. The task loss weights are all set as 1. 

\begin{table}[!htp]
\centering
\renewcommand{\arraystretch}{1.5}
\begin{small}
    \begin{tabular}{lcccccc}
        \specialrule{1.1pt}{0pt}{1pt}
        Tasks &Gen-T          &Gen-TS         &Gen-R          &Refinement     &Completion      &UGen\\
        Weights  &$\frac{1}{12}$ &$\frac{1}{3}$  &$\frac{1}{12}$ &$\frac{1}{3}$  &$\frac{1}{12}$    &$\frac{1}{12}$\\
        \specialrule{1.1pt}{2pt}{0pt}
    \end{tabular}
\end{small}
\caption{Task sample weights in constructing mixed dataset for LayoutFormer++(M).}
\label{Tab:sample_weight}
\end{table}

The quantitative comparison between LayoutFormer++(M) with the state-of-the-art baselines of each task are shown in Table~\ref{tab:mixed_tasks}. We highlight the results of LayoutFormer++(M) by bold when they achieve better performance than the baselines. We find that although handling all the tasks simultaneously is much more difficult than tackling one task, our LayoutFormer++(M) still significantly outperforms the baselines on most metrics. This indicates that our LayoutFormer++ can be trained as an all-in-one model, to flexibly handle different layout generation tasks by a single set of model parameters.

\begin{table*}[ht]
    \centering
    \renewcommand{\arraystretch}{1.3}
        \resizebox{\textwidth}{!}{
            \begin{tabular}{llcccccccc}
                \specialrule{1.1pt}{0pt}{1pt}
                                            &                   & \multicolumn{4}{c}{RICO}   & \multicolumn{4}{c}{PubLayNet}    \\ 
                                            \cmidrule(l){3-6} \cmidrule(l){7-10}
                Tasks                       & \makecell{Methods}& mIoU $\uparrow$   & FID $\downarrow$  & Align. $\downarrow$   & Overlap $\downarrow$  
                                                                & mIoU $\uparrow$   & FID $\downarrow$  & Align. $\downarrow$   & Overlap $\downarrow$      \\ \hline
                \multirow{2}{*}{Gen-T}      & LayoutGAN++       & 0.298             & 5.954             & 0.261                 & 0.620                 
                                                                & 0.297             & 14.875            & 0.124                 & 0.148                     \\ 
                                            & LayoutFormer++(M) & \textbf{0.396}    & \textbf{2.101}    & \textbf{0.161}        & \textbf{0.586}                      
                                                                & \textbf{0.352}    & \textbf{10.620}   & \textbf{0.021}        & \textbf{0.018}            \\ % 
                                            \hline

                \multirow{2}{*}{Gen-TS}     & BLT               & 0.604             & 0.951             & 0.181                 & 0.660                          
                                                                & 0.428             & 7.914             & 0.021                 & 0.419                                   \\ 
                                            & LayoutFormer++(M) & 0.577             & 1.392             & \textbf{0.179}        & \textbf{0.567}           
                                                                & \textbf{0.463}    & \textbf{2.097}    & 0.026                 & \textbf{0.041}            \\ % 
                                            \hline
                \multirow{2}{*}{Gen-R}      & CLG-LO            & 0.286             & 8.898             & 0.311                 & 0.615                 
                                                                & 0.277             & 19.738            & 0.123                 & 0.200                       \\  
                                            & LayoutFormer++(M) & \textbf{0.372}    & 11.026            & \textbf{0.122}        & \textbf{0.593}                     
                                                                & \textbf{0.318}    & \textbf{11.694}   & \textbf{0.024}       & \textbf{0.123}           \\ % 
                                            \hline
                \multirow{2}{*}{Refinement} & RUITE             & 0.811             & 0.107             & 0.133                 & 0.483                  
                                                                & 0.781             & 0.061             & 0.029                 & 0.020                 \\
                                            & LayoutFormer++(M) & 0.786             & \textbf{0.084}    & 0.135                 & 0.495                 
                                                                & 0.773             & 0.094             & \textbf{0.022}        & \textbf{0.006}        \\
                                            \hline
                \multirow{2}{*}{Completion} & LayoutTransformer & 0.363             & 6.679             & 0.194                 & 0.478                 
                                                                & 0.077             & 14.769            & 0.019                 & 0.0013                \\
                                            & LayoutFormer++(M) & \textbf{0.731}    & \textbf{4.104}    & \textbf{0.074}        & \textbf{0.472}                 
                                                                & \textbf{0.475}    & \textbf{8.304}    & 0.023                 & 0.0016                \\
                                            \hline
                \multirow{2}{*}{UGen}       & LayoutTransformer & 0.439             & 22.884            & 0.052                 & 0.471                  
                                                                & 0.062             & 36.304            & 0.031                 & 0.0009                \\
                                            & LayoutFormer++(M) & \textbf{0.734}    & \textbf{11.667}   & 0.058                 & \textbf{0.463}                  
                                                                & \textbf{0.430}    & \textbf{30.161}   & \textbf{0.029}        & \textbf{0.0008}        \\
                \specialrule{1.1pt}{1pt}{0pt}
            \end{tabular}}
    \caption{Quantitative comparisons between LayoutFormer++(M) and task-specific baselines on six tasks and two datasets.}
    \label{tab:mixed_tasks}
\end{table*}

\subsection{Generalization to New Tasks}
\label{sec:combined_tasks}
In this section, we aim to demonstrate that LayoutFormer++ can be flexibly adapted to the new layout generation tasks. We propose four new tasks to develop the experiments: \emph{Gen-TC}, \emph{Gen-TSC}, \emph{Gen-RS} and \emph{Gen-RP}, by combining the existing typical tasks:

\emph{\textbf{Gen-TC}} is the combination of Gen-T and Completion.
In this task, user has already placed some elements on the layout, while specifying the types of other elements that are required to be arranged. 
% For the placed layout, since their attributes are completely given by the user, we denote them as complete elements in the following.
We formulate the input as $S_{\text{Gen-TC}}=\{\langle\text{sos}\rangle  c_1 x_1 y_1 w_1 h_1 | \dots | c_P x_P y_P w_P h_P || c_{P+1} | \dots | c_N  \langle\text{eos}\rangle\}$, where $P$ is the number of already placed elements, and $N$ denotes the total number of elements.

\emph{\textbf{Gen-TSC}} is the combination of Gen-TS and Completion. 
In this tasks, user has already placed some elements on the layout, while specifying the types and sizes of other elements that are required to be arranged. 
We formulate the input as $S_{\text{Gen-TSC}}=\{\langle\text{sos}\rangle  c_1 x_1 y_1 w_1 h_1 | \dots | c_P x_P y_P w_P h_P || c_{P+1} w_{P+1} h_{P+1} | \dots | c_N w_N h_N \langle\text{eos}\rangle\}$.

\emph{\textbf{Gen-RS}} combines the Gen-R with Gen-TS. 
In this task, user specifies the types and sizes of the elements, and also require the relative position relationships (i.e., above, bottom, left, right, and overlap) between some elements. 
The input constraint sequence is formulated as $S_{\text{Gen-RS}}=\{\langle\text{sos}\rangle  c_1 w_1 h_1| c_2 w_2 h_2| \dots | c_N w_N h_N|| c_{k_1} k_1 r_{k_1, k_2} c_{k_2} k_2 | \dots  | c_{k_{2M-1}} k_{2M-1} r_{k_{2M-1}, k_{2M}} \\ c_{k_{2M}} k_{2M} \langle\text{eos}\rangle\}$, where $M$ is the number of the positional relationships.

\emph{\textbf{Gen-RP}} generates layouts from user-specified element types and positions, and relative size relationships (i.e., smaller, larger and equal) between the elements. The input constraint sequence is formulated as $S_{\text{Gen-RP}}=\{\langle\text{sos}\rangle  c_1 x_1 y_1| c_2 x_2 y_2| \dots |\\ c_N x_N y_N|| c_{k_1} k_1 r_{k_1, k_2} c_{k_2} k_2 | \dots  | c_{k_{2M-1}} k_{2M-1} r_{k_{2M-1}, k_{2M}} c_{k_{2M}} k_{2M} \langle\text{eos}\rangle\}$, where $M$ is the number of the size relationships.

We leverage LayoutFormer++ for the four new tasks on the Rico dataset. For Gen-TC and Gen-TSC, we random sample 50\% elements of each layout as the given complete elements. For Gen-RS and Gen-RP, same as in Gen-R, we randomly sample $10\%$ element relationships as the input. 

Table~\ref{tab:combined_tasks} shows the quantitative results of these combined new tasks. Since there is no baseline for the new tasks, we just give the results of the related typical tasks at the bottom of Table~\ref{tab:combined_tasks} for reference. 
Figure~\ref{fig:new_tasks} shows the qualitative results for the combined tasks. For each task, we show three groups of layouts, where the right one in each group is the generation by LayoutFormer++, and the left one is the real layout where the input constraints for inference come from.
It can be obtained that LayoutFormer++ can flexibly handle all these new layout generation tasks with achieving good generation quality.

\begin{table*}[ht]
    \centering
    \renewcommand{\arraystretch}{1.1}
            \begin{tabular}{llcccc}
                \specialrule{1.1pt}{0pt}{1pt}
                & \multicolumn{1}{l}{Tasks}                       & mIoU $\uparrow$   & FID $\downarrow$  & Align. $\downarrow$   & Overlap $\downarrow$      \\ \hline
                \multirow{4}{*}{Combined Tasks}
                & \multirow{1}{*}{Gen-TC}      & 0.744             & 0.862             & 0.093                  & 0.562 \\ 
                                            
                & \multirow{1}{*}{Gen-TSC}     & 0.818             & 0.534             & 0.129                 & 0.529\\ % 
                                            
                & \multirow{1}{*}{Gen-RS}      & 0.593            & 5.022             & 0.190                  & 0.561 \\ % 
                                            
                & \multirow{1}{*}{Gen-RP}     & 0.799             & 0.769             & 0.119                 & 0.550 \\
                \hline
                \multirow{3}{*}{Typical Tasks}
                & \multirow{1}{*}{Gen-T}      & 0.432             & 1.096             & 0.230                  & 0.530 \\ 
                                            
                & \multirow{1}{*}{Gen-TS}     & 0.620             & 0.757             & 0.202                 & 0.542\\ % 
                                            
                & \multirow{1}{*}{Gen-R}      & 0.424            & 5.972             & 0.332                  & 0.537 \\ % 
                \specialrule{1.1pt}{1pt}{0pt}
            \end{tabular}
    \caption{Quantitative evaluation for the combined tasks on Rico.}
    \label{tab:combined_tasks}
\end{table*}

\begin{figure}[!htbp]
    \centering
    \begin{minipage} {\linewidth}
    \begin{subfigure}[Gen-TC]{\linewidth}
        \includegraphics[width=\linewidth]{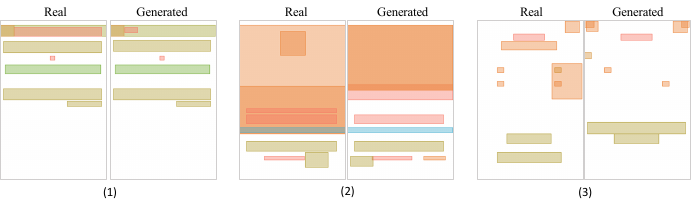}
        \caption{Gen-TC}
    \end{subfigure} %
    \end{minipage}
    
    \medskip
    \caption{Qualitative results of Gen-TC, Gen-TSC, Gen-RS and Gen-RP.}
    
\end{figure}

\begin{figure}[!htbp]\ContinuedFloat
    \centering   
    
    \begin{minipage} {\linewidth}
    \begin{subfigure}[Gen-TSC]{\linewidth}

        \includegraphics[width=\linewidth]{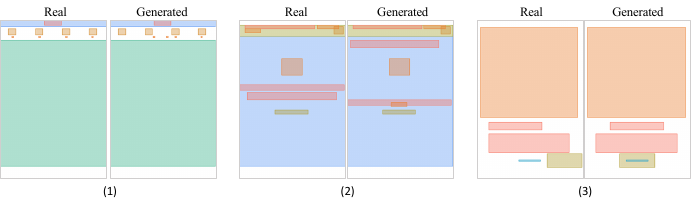}
        \caption{Gen-TSC}
    \end{subfigure} %
    \end{minipage}
    
    \medskip

    \begin{minipage} {\linewidth}
    \begin{subfigure}[Gen-RS]{\linewidth}

        \includegraphics[width=\linewidth]{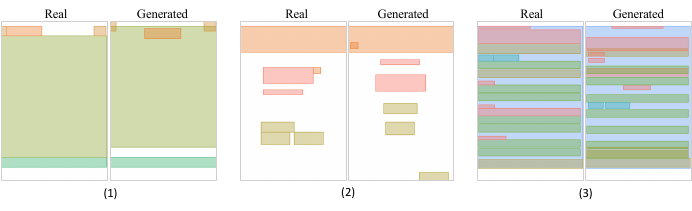}
        \caption{Gen-RS}
    \end{subfigure} %
    \end{minipage}
    
    \medskip
    
    \begin{minipage}{\linewidth}
    \begin{subfigure}[Gen-RP]{\linewidth}   

        \includegraphics[width=\linewidth]{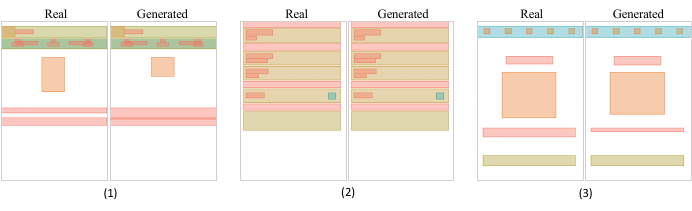}
        \caption{Gen-RP}
    \end{subfigure} 
    \end{minipage}
    \caption{ (Cont.) Qualitative results of Gen-TC, Gen-TSC, Gen-RS and Gen-RP.}
    \label{fig:new_tasks}
\end{figure}

\newpage
\section{More Discussion on Decoding Space Restriction}
\label{sec:compare_with_topk}

As we introduced in the paper, the decoding space restriction strategy prunes the predicted distribution at each decoding step, and restarts the decoding process from a previous step when current prediction cannot achieve good controllability. 
Another naive and straightforward approach is running the decoding process multiple times and then selecting the best generation among them as the final results. In this section, we compare the decoding space restriction strategy with this naive approach.

We develop the comparison on the Gen-R tasks. 
For the naive approach, we run the decoding process for 5 and 3 times on Rico and PubLayNet datasets respectively, which is same as the number of the max back times we set for the backtracking mechanism. 
We try different metrics to select the result from multi-times decoding for evaluation. \emph{select by Align.} denotes that for each layout, the result with the minimum value of align. is chosen for developing the evaluation. \emph{select by Overlap} denotes the result is selected by the overlap value, and \emph{select by Vio.\%} denotes the results is selected by the Vio.\%.

Table~\ref{tab:compare_topk} shows the quantitative comparison. We have following observations.
For \emph{select by Align.} and \emph{select by Overlap}, they significantly improve the performance on Align. and Overlap respectively. However, they do not perform well on Vio \%. 
\emph{select by  Vio.\%} achieves better Vio \% than \emph{select by Align.} and \emph{select by Overlap}.
However, LayoutFormer++ significantly outperforms \emph{select by  Vio.\%} on Vio.\%, while achieving comparable performance on the quality metrics with all the baselines.
This demonstrates the advantage of the decoding space restriction strategy.
Without the restriction to the predicted distribution, the naive approach is very difficult to sample the attributes for the layout to conform all the constraints, even sample multiple times. 

\begin{table*}[ht]
    \centering
    \renewcommand{\arraystretch}{1.1}
        \resizebox{0.95\textwidth}{!}{
            \begin{tabular}{llcccccccccc}
                \specialrule{1.1pt}{0pt}{1pt}
                                            &                   & \multicolumn{5}{c}{RICO}   & \multicolumn{5}{c}{PubLayNet}    \\ 
                                            \cmidrule(l){3-7} \cmidrule(l){8-12}
                Tasks                       & \makecell{Methods}& mIoU $\uparrow$   & FID $\downarrow$  & Align. $\downarrow$   & Overlap $\downarrow$  & Vio. \% $\downarrow$
                                                                & mIoU $\uparrow$   & FID $\downarrow$  & Align. $\downarrow$   & Overlap $\downarrow$  & Vio. \% $\downarrow$    \\ \hline
                \multirow{4}{*}{Gen-R}      & select by Align.  & 0.454             & 5.708             & 0.096                 & 0.560                 & 32.97               
                                                                & 0.359             & 4.640             & 0.0007                & 0.029                 & 16.05   \\ 
                                            & select by Overlap & 0.454             & 6.466             & 0.336                 & 0.422                 & 32.66
                                                                & 0.352             & 5.397             & 0.023                 & 0.0003                & 15.74   \\ 
                                            & select by Vio.\%  & 0.471             & 4.552             & 0.316                 & 0.542                 & 27.45             
                                                                & 0.361             & 4.344             & 0.022                 & 0.027                 & 6.6   \\  
                                            & LayoutFormer++    & 0.424             & 5.972             & 0.332                 & 0.537                 & \textbf{11.84}            
                                                                & 0.353             & 4.954             & 0.025                 & 0.076                 & \textbf{3.9}   \\ % 
                \specialrule{1.1pt}{1pt}{0pt}
            \end{tabular}}
    \caption{Quantitative comparisons with the naive approach.}
    \label{tab:compare_topk}
\end{table*}

% \newpage
\section{More Qualitative Results}
\label{sec:qualitative_supp}

We show more generated layouts here to better demonstrate the good performance of LayoutFormer++.
For all the six typical layout generation tasks, we present the generated layouts by LayoutFormer++ on both RICO and PubLayNet.

\textbf{Gen-T. }
The generated layouts of Gen-T are shown in Figure~\ref{fig:gen-t}. We show six groups of generated results on both Rico and PubLayNet, each group contains two layouts generated from the same element type constraints below. 
% It can be seen that LayoutFormer++ can generate aesthetics and diverse layouts while satisfying the same constraints.

\textbf{Gen-TS. }
The generated layouts of Gen-TS on Rico and PubLayNet are shown in Figure~\ref{fig:gen-ts}. Similar with the results of Gen-T, each group contains two layouts generated from the same element type and size constraints in the below table.

\textbf{Gen-R. }
The generated layouts of Gen-R on Rico and PubLayNet are shown in Figure~\ref{fig:gen-r}. We list the relationship constraints that the layouts generated from in the below table.

\begin{figure}[!htbp]
    \centering
    \begin{minipage} {\linewidth}
    \begin{subfigure}[Rico]{\linewidth}

        \includegraphics[width=\linewidth]{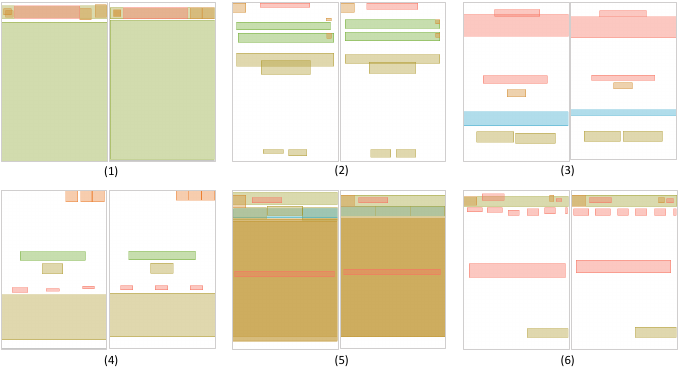}
        \caption{Rico}
        % \label{fig:refinement-rico}
    \end{subfigure} %
    \end{minipage}
    \medskip

    \begin{minipage}{\linewidth}
    \begin{subfigure}[PubLayNet]{\linewidth}   

        \includegraphics[width=\linewidth]{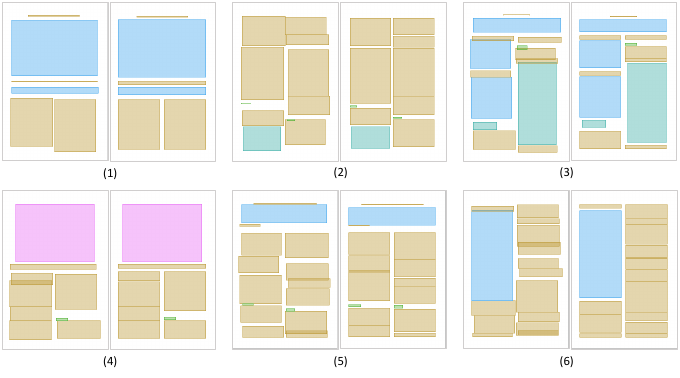}
        \caption{PubLayNet}
        % \label{fig:refinement-pln}    
    \end{subfigure} 
    \end{minipage}
    \caption{Qualitative results of Refinement on Rico and PubLayNet.}
    \label{fig:refinement}
\end{figure}

\textbf{Refinement. }
The generated layouts of Refinement on Rico and PubLayNet are shown in Figure~\ref{fig:refinement}. We show six groups of generated layouts, where each group contains two layouts. The left one is the noised layout that needs refinement, and the right one is the layout refined by LayoutFormer++.

\begin{figure}[!htbp]
    \centering
    \begin{minipage} {\linewidth}
    \begin{subfigure}[Rico]{\linewidth}

        \includegraphics[width=\linewidth]{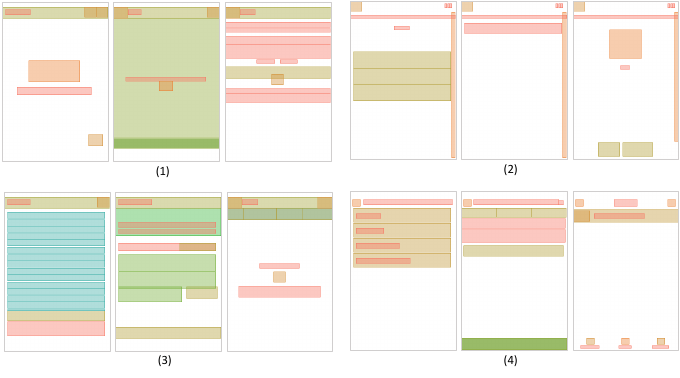}
        \caption{Rico}
        % \label{fig:completion-pln}
    \end{subfigure} %
    \end{minipage}
    \medskip

    \begin{minipage}{\linewidth}
    \begin{subfigure}[PubLayNet]{\linewidth}   

        \includegraphics[width=\linewidth]{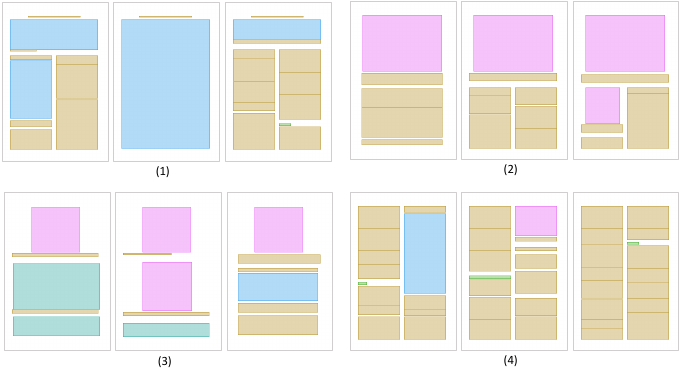}
        \caption{PubLayNet}
        % \label{fig:completion-pln}    
    \end{subfigure} 
    \end{minipage}
    \caption{Qualitative results of Completion on Rico and PubLayNet.}
    \label{fig:completion}
\end{figure}

\textbf{Completion. }
The generated layouts of Completion are shown in Figure~\ref{fig:completion}. We show four groups of generated layouts on both Rico and PubLayNet, where each group contains three layouts generated from the same first element. By leveraging top-k sampling, LayoutFormer++ can complete the same partial layout into various final layouts.

\begin{figure}[!htbp]
    \centering
    \begin{minipage} {\linewidth}
    \begin{subfigure}[Rico]{\linewidth}

        \includegraphics[width=\linewidth]{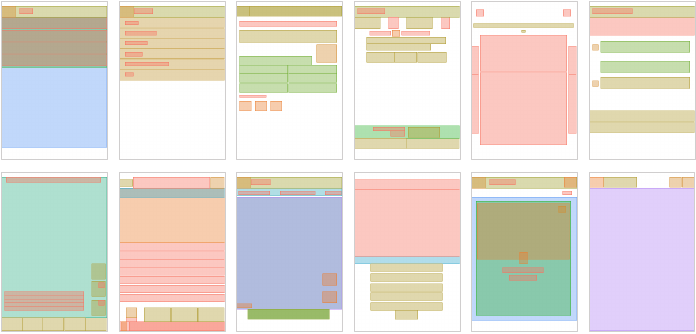}
        \caption{Rico}
        % \label{fig:ugen-pln}
    \end{subfigure} %
    \end{minipage}
    \medskip

    \begin{minipage}{\linewidth}
    \begin{subfigure}[PubLayNet]{\linewidth}   

        \includegraphics[width=\linewidth]{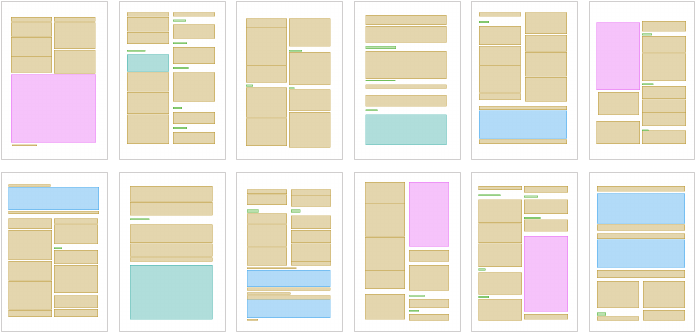}
        \caption{PubLayNet}
        % \label{fig:ugen-pln}    
    \end{subfigure} 
    \end{minipage}
    \caption{Qualitative results of UGen on Rico and PubLayNet.}
    \label{fig:ugen}
\end{figure}

\textbf{UGen. }
The generated layouts of UGen on RICO and PubLayNet are shown in Figure~\ref{fig:ugen}. The top-k sampling ensures the diversity of generated layouts from identical input for LayoutFormer++.

\begin{figure}[!htbp]
    \centering
    \begin{minipage} {\linewidth}
    \begin{subfigure}[Rico]{\linewidth}

        \includegraphics[width=\linewidth]{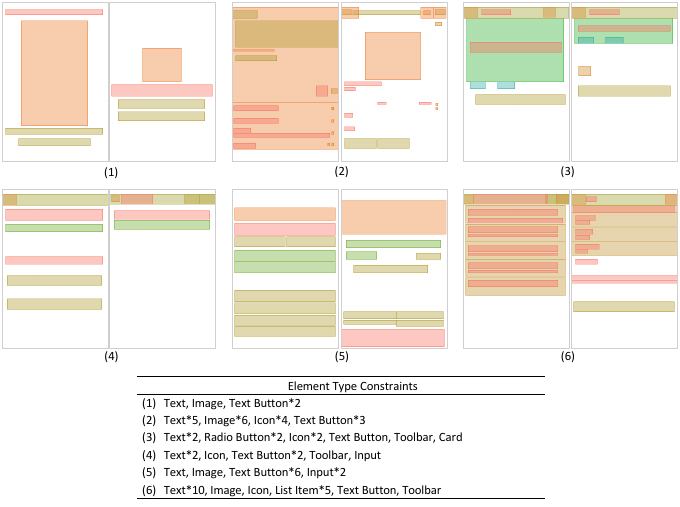}
        \caption{Rico}
        % \label{fig:gen-t-rico}
    \end{subfigure} %
    \end{minipage}
    \caption{Qualitative results of Gen-T on Rico and PubLayNet. The element type constraints are in the table.}
\end{figure}

\begin{figure}[!htbp]\ContinuedFloat
    \centering
    \begin{minipage}{\linewidth}
    \begin{subfigure}[PubLayNet]{\linewidth}   

        \includegraphics[width=\linewidth]{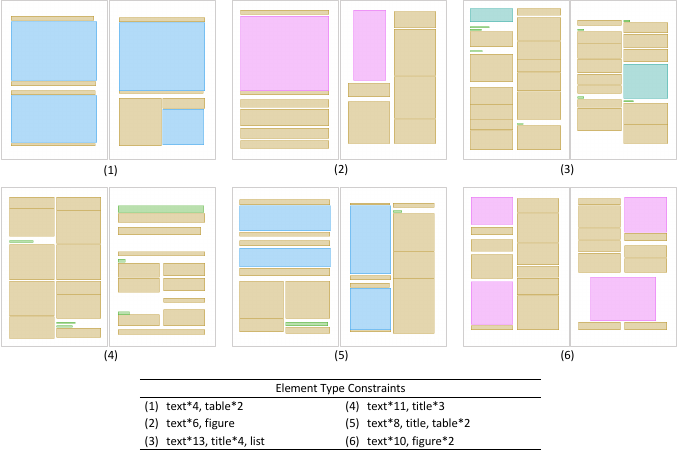}
        \caption{PubLayNet}
        % \label{fig:gen-t-pln}    
    \end{subfigure} 
    \end{minipage}
    \caption{(Cont.) Qualitative results of Gen-T on Rico and PubLayNet. The element type constraints are in the table.}
    \label{fig:gen-t}
\end{figure}

\begin{figure}[!htbp]
    \centering
    \begin{minipage} {\linewidth}
    \begin{subfigure}[Rico]{\linewidth}

        \includegraphics[width=\linewidth]{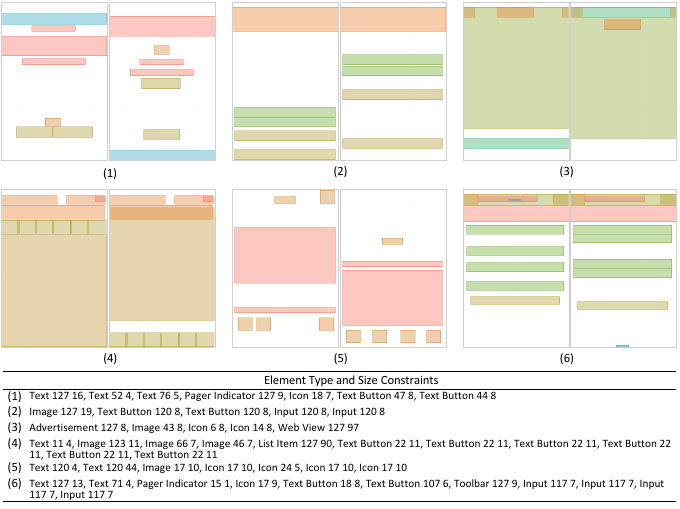}
        \caption{Rico}
        % \label{fig:gen-ts-rico}
    \end{subfigure} %
    \end{minipage}
    \medskip
    \caption{Qualitative results of Gen-TS on Rico and PubLayNet. The tables show the element type and size constraints.}
\end{figure}

\begin{figure}[!htbp]\ContinuedFloat
    \centering
    \begin{minipage}{\linewidth}
    \begin{subfigure}[PubLayNet]{\linewidth}   

        \includegraphics[width=\linewidth]{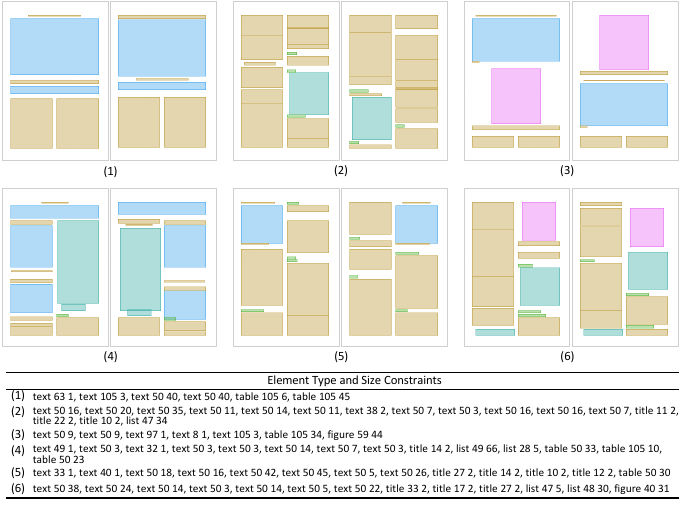}
        \caption{PubLayNet}
        % \label{fig:gen-ts-pln}    
    \end{subfigure} 
    \end{minipage}
    \caption{(Cont.) Qualitative results of Gen-TS on Rico and PubLayNet. The tables show the element type and size constraints.}
    \label{fig:gen-ts}
\end{figure}

\begin{figure}[!htbp]
    \centering
    \begin{minipage} {\linewidth}
    \begin{subfigure}[Rico]{\linewidth}

        \includegraphics[width=\linewidth]{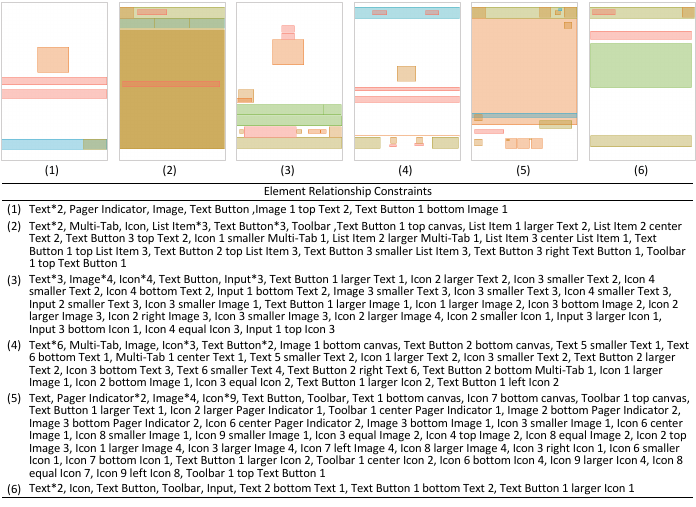}
        \caption{Rico}
        % \label{fig:gen-r-rico}
    \end{subfigure} %
    \end{minipage}
    \caption{Qualitative results of Gen-R on Rico and PubLayNet. The tables show the element relationship constraints.}
\end{figure}

\begin{figure}[!htbp]\ContinuedFloat
    \centering
    \begin{minipage}{\linewidth}
    \begin{subfigure}[PubLayNet]{\linewidth}   

        \includegraphics[width=\linewidth]{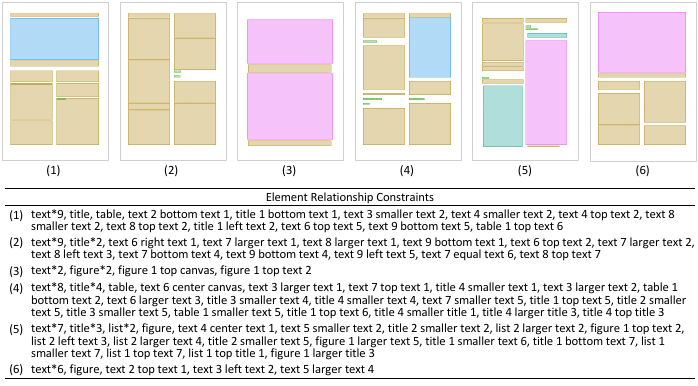}
        \caption{PubLayNet}
        % \label{fig:gen-r-pln}    
    \end{subfigure} 
    \end{minipage}
    \caption{(Cont.) Qualitative results of Gen-R on Rico and PubLayNet. The tables show the element relationship constraints.}
    \label{fig:gen-r}
\end{figure}

{\small
\bibliographystyle{ieee_fullname}
\bibliography{supplementary}
}